\documentclass[final]{cvpr}

\usepackage{times}
\usepackage{epsfig}
\usepackage{graphicx}
\usepackage{amsmath}
\usepackage{amssymb}
\usepackage{booktabs}
\usepackage{multirow}

\usepackage[pagebackref=true,breaklinks=true,colorlinks,bookmarks=false]{hyperref}

\def\cvprPaperID{****} % *** Enter the CVPR Paper ID here
\def\confYear{CVPR 2022}

\begin{document}

%%%%%%%%% TITLE
\title{NTIRE 2022 Challenge on Perceptual Image Quality Assessment}

\author{
% organizers
Jinjin Gu \and Haoming Cai \and Chao Dong \and Jimmy S. Ren \and Radu Timofte \and
% track 1 team ranked 1
Yuan Gong \and Shanshan Lao \and Shuwei Shi \and Jiahao Wang \and Sidi Yang \and Tianhe Wu \and Weihao Xia \and Yujiu Yang \and
% track 2 team ranked 1
Mingdeng Cao \and  % There are also authors who duplicate the track 1 team ranked 1
% track 1 team ranked 2
Cong Heng \and Lingzhi Fu \and Rongyu Zhang \and Yusheng Zhang \and Hao Wang \and Hongjian Song \and
% track 2 team ranked 2
Jing Wang \and Haotian Fan \and Xiaoxia Hou \and
% track 1 team ranked 3
Ming Sun \and Mading Li \and Kai Zhao \and Kun Yuan \and Zishang Kong \and Mingda Wu \and Chuanchuan Zheng \and
% track 2 team ranked 3
Marcos V. Conde \and Maxime Burchi \and
% track 1 team ranked 4
% There are also authors who duplicate the track 2 team ranked 3
% track 2 team ranked 4
% There are also authors who duplicate the track 1 team ranked 3
% track 1 team ranked 5
Longtao Feng \and Tao Zhang \and Yang Li \and Jingwen Xu \and Haiqiang Wang \and Yiting Liao \and Junlin Li \and
% track 2 team ranked 5
% There are also authors who duplicate the track 1 team ranked 2
% track 1 team ranked 6
Kele Xu \and Tao Sun \and Yunsheng Xiong \and
% track 2 team ranked 6
% Minsu Kwon \and
% track 1 team ranked 7
Abhisek Keshari \and Komal \and Sadbhawana Thakur \and Vinit Jakhetiya \and Badri N Subudhi \and
% track 2 team ranked 7
Hao-Hsiang Yang \and Hua-En Chang \and Zhi-Kai Huang \and Wei-Ting Chen \and Sy-Yen Kuo \and
% track 1 team ranked 8
Saikat Dutta \and Sourya Dipta Das \and Nisarg A. Shah \and Anil Kumar Tiwari
}

\maketitle

%%%%%%%%% ABSTRACT
\begin{abstract}
This paper reports on the NTIRE 2022 challenge on perceptual image quality assessment (IQA), held in conjunction with the New Trends in Image Restoration and Enhancement workshop (NTIRE) workshop at CVPR 2022.
This challenge is held to address the emerging challenge of IQA by perceptual image processing algorithms.
The output images of these algorithms have completely different characteristics from traditional distortions and are included in the PIPAL dataset used in this challenge.
This challenge is divided into two tracks, a full-reference IQA track similar to the previous NTIRE IQA challenge and a new track that focuses on the no-reference IQA methods.
The challenge has 192 and 179 registered participants for two tracks.
In the final testing stage, 7 and 8 participating teams submitted their models and fact sheets.
Almost all of them have achieved better results than existing IQA methods, and the winning method can demonstrate state-of-the-art performance.
\end{abstract}
{\let\thefootnote\relax\footnotetext{%
\hspace{-5mm}$^*$Jinjin Gu (\texttt{jinjin.gu@sydney.edu.au}), Haoming Cai, Chao Dong, Jimmy Ren and Radu Timofte are the NTIRE 2022 challenge organizers. The other authors participated in the challenge.
Appendix.~\ref{sec:apd:track1team} and Appendix.~\ref{sec:apd:track2team} contain the authors' team names and affiliations. The NTIRE website: \url{https://data.vision.ee.ethz.ch/cvl/ntire22/}}}

%%%%%%%%% BODY TEXT
\section{Introduction}
Assessing the perceptual quality of an image is a fundamental requirement in the fields of image acquisition, transmission, compression, reproduction, and processing.
Image quality assessment (IQA) methods are tools that use computational models to measure the perceptual quality of images.
As the ``evaluation mechanism'', IQA plays a critical role in guiding the development of image processing algorithms.
However, distinguish perceptually better images is not an easy task \cite{tid2013,pipal,gu2020image}, especially as newly-appeared image distortion types continue to challenge IQA methods, \eg, Generative Adversarial Networks (GANs) based algorithms \cite{goodfellow2014generative} and perceptual-oriented algorithms \cite{johnson2016perceptual,srgan2017,wang2018esrgan,zhang2019ranksrgan}.
IQA methods that are capable of automatically and accurately predicting subjective quality are in demand nowadays.

The NTIRE 2022 Perceptual Image Quality Assessment Challenge aims to push the developing state-of-the-art perceptual image quality assessment methods to deal with the novel GAN-based distortion types and gain new insights.
We employ the PIPAL dataset \cite{pipal} in this challenge, which is the only dataset including the results of perceptual-oriented algorithms.
The PIPAL dataset contains 250 reference images, 29k distorted images and 1.88 million human judgements.
The large size and diversity of distortion types of the PIPAL dataset allow us to benchmark these IQA methods.

This is the second perceptual IQA challenge held at the NTIRE workshop.
In the last challenge \cite{gu2021ntire}, several submitted entries significantly outperformed existing methods and achieved state-of-the-art performance in the full-reference IQA field.
In this challenge, we include two tracks.
The first track is similar to the NTIRE 2021 IQA challenge, focusing on full-reference methods.
Considering the wide range of application scenarios and demands of no-reference methods, we set up a second track that focuses on no-reference IQA methods.
We anticipate this new track to push developing state-of-the-art no-reference IQA methods.

The challenge has 192 and 179 registered participants for two tracks, respectively.
Among them, 7 and 8 participating teams submitted their models and fact sheets in the final testing stage, respectively.
They introduce new technologies in network architectures, loss functions, ensemble methods, data augmentation methods, and \etc.
We present detailed challenge results in Sec~\ref{sec:results}.

This challenge is one of the NTIRE 2022 associated challenges: spectral recovery~\cite{arad2022ntirerecovery}, spectral demosaicing~\cite{arad2022ntiredemosaicing},
perceptual image quality assessment~\cite{gu2022ntire},
inpainting~\cite{romero2022ntire},
night photography rendering~\cite{ershov2022ntire},
efficient super-resolution~\cite{li2022ntire},
learning the super-resolution space~\cite{lugmayr2022ntire},
super-resolution and quality enhancement of compressed video~\cite{yang2022ntire},
high dynamic range~\cite{perezpellitero2022ntire},
stereo super-resolution~\cite{wang2022ntire},
burst super-resolution~\cite{bhat2022ntire}.

\begin{table*}[t]
    \centering
    \caption{Quantitative results for the NTIRE 2022 Perceptual IQA challenge.}
    \label{tab:main_results}

        \begin{tabular}{c|p{4cm}|p{4cm}|ccc}
        \toprule
        \multirow{2}{*}{Rank} & \multirow{2}{*}{Team Name} & \multirow{2}{*}{Author/Method} & \multicolumn{3}{c}{PIPAL-NTIRE22-Test}\\
        & & & Main Score & SRCC & PLCC\\
        \midrule
        \multicolumn{6}{c}{Track 1: Full-Reference IQA}\\
        \midrule
        1 & THU1919Group      & shanshan    & 1.6511 & 0.8227 & 0.8284\\
        2 & Netease OPDAI     & CongHeng.   & 1.6422 & 0.8152 & 0.8271\\
        3 & KS                & JustTryTry  & 1.6404 & 0.8170 & 0.8235\\
        4 & JMU-CVLab         & burchim     & 1.5406 & 0.7659 & 0.7747\\
        5 & Yahaha!           & FLT         & 1.5375 & 0.7654 & 0.7722\\
        6 & debut\_kele       & debut       & 1.5006 & 0.7372 & 0.7634\\
        7 & Pico Zen          & Komal       & 1.4504 & 0.7129 & 0.7375\\
        8 & Team Horizon      & tensorcat   & 1.4032 & 0.7006 & 0.7027\\
        \midrule
        \multirow{6}{*}{} & \multirow{6}{*}{Baselines} 
        & IQT (NTIRE-21 Winner)    & 1.5884 & 0.7895 & 0.7989 \\
        & & LPIPS-Alex            & 1.1369 & 0.5658 & 0.5711 \\
        & & LPIPS-VGG             & 1.2278 & 0.5947 & 0.6331 \\
        & & DISTS                 & 1.3422 & 0.6548 & 0.6873 \\
        & & SSIM                  & 0.7530 & 0.3615 & 0.3915 \\
        & & PSNR                  & 0.5263 & 0.2493 & 0.2769 \\
        \midrule
        \multicolumn{6}{c}{Track 2: No-Reference IQA}\\
        \midrule
        1 & THU\_IIGROUP       & THU\_IIGROUP & 1.4436 & 0.7040 & 0.7396 \\
        2 & DTIQA             & EvaLab.     & 1.4367 & 0.6996 & 0.7371 \\
        3 & JMU-CVLab         & nanashi     & 1.4219 & 0.6965 & 0.7254 \\
        4 & KS                & JustTryTry  & 1.4066 & 0.6808 & 0.7257 \\
        5 & NetEase OPDAI     & wanghao1003 & 1.3902 & 0.6705 & 0.7196 \\
        6 & Withdrawn submission        & anonymous   & 1.1828 & 0.5760 & 0.6068 \\
        7 & NTU607QCO-IQA     & mrchang87   & 1.1117 & 0.5269 & 0.5848 \\
        \midrule
        \multirow{4}{*}{} & \multirow{4}{*}{Baselines} 
        & NIQE         & 0.1418 & 0.0300 & 0.1118 \\
        & & MA         & 0.3978 & 0.1737 & 0.2242 \\
        & & PI         & 0.2764 & 0.1234 & 0.1529 \\
        & & Brisque    & 0.5722 & 0.2695 & 0.3027 \\
        % & & LPIPS-Alex & 1.1755 & 0.5839 & 0.5916 \\
        \bottomrule
    \end{tabular}
\end{table*}

%-------------------------------------------------------------------------
\section{Related Work}
\paragraph{Full-Reference Image quality assessment (FR-IQA).}
FR-IQA methods evaluate the similarity between a distorted image and a given reference image and have been widely used to evaluate image/video processing algorithms.
FR-IQA methods follow a long line of works, the most well-known of which is PSNR and SSIM \cite{ssim}.
SSIM introduces structural information in measuring image similarity and opens a precedent for evaluating image structure or feature similarity.
After that, various FR-IQA methods have been proposed to bridge the gap between the results of IQA methods and human judgements \cite{ms-ssim,sr-sim,fsim,ifc,vsi}.
Similar to other computer vision problems, advanced data-driven methods have also motivated the investigation of applications of IQA, such as LPIPS \cite{zhang2018unreasonable}, PieAPP \cite{prashnani2018pieapp}, WaDIQaM \cite{wadiqam}, SDW \cite{gu2020image} and DISTS \cite{dists}.
The 2021 NTIRE challenge has also brought some excellent FR-IQA methods, i.e., Cheon \etal \cite{IQT2021ntire} propose a transformer-based FR-IQA method IQT and win the first place at the challenge, Guo \etal \cite{IQMA2021ntire} propose bilateral-branch multi-scale image quality estimation (IQMA) network, and Shi \etal \cite{RADN2021ntire} propose Region Adaptive Deformable Network (RADN).

\paragraph{No-Reference Image quality assessment (NR-IQA).}
In addition to the above FR-IQA methods, NR-IQA methods are proposed to assess image quality without a reference image.
A typical NR-IQA is often based on natural image statistics.
Natural images usually follow these natural image prior distributions, while distorted images often break such statistical regularities.
Variation of methods have been used to extract natural image statistics \cite{pan2022no,mittal2012making,zhang2015feature,saad2012blind,moorthy2011blind,zhang2015som,ye2012unsupervised}.
In the era of deep learning, deep networks are anticipated to replace hand-crafted feature extraction and learn statistical priors on images, and many deep learning-based NR-IQA methods are proposed \cite{kang2014convolutional,bosse2017deep,lin2018hallucinated,talebi2018nima,zhang2018blind,bianco2018use,yan2019naturalness,su2020blindly,zhu2020metaiqa,zhang2020learning}.
More related to this work, Blau \etal \cite{blau2018perception} combine two NR-IQA methods, Ma \cite{ma2017learning} and NIQE \cite{niqe} and propose the Perceptual Index (PI) method to measure the perceptual quality of super-resolution results without reference image.
Although it can lead to the development of better perceptual-oriented algorithms compared with other FR-IQA methods that focus on evaluating distortion, its IQA performance is still unsatisfactory.
In this challenge, we set a new track that focuses on NR-IQA methods and bring more advanced NR-IQA methods to this field.

\paragraph{Perceptual-oriented and GAN-based distortion.}
In the past years, benefiting from the invention of perceptual-oriented loss function \cite{johnson2016perceptual,wang2018esrgan} and GANs \cite{goodfellow2014generative}. many photo-realistic image generation and processing algorithms are proposed \cite{srgan2017,wang2018esrgan,sftgan2018,zhang2019ranksrgan,Cai2021CUGAN}.
These perceptual-oriented algorithms greatly improve the perceptual effect of the output image.
However, they also bring completely new characteristics to the output images.
In general, these methods often fabricate seemingly realistic yet fake details and textures.
They do not quite match the quality of detail loss, as they usually contain texture-like noise, or the quality of noise, the noise is similar to the ground truth in appearance but is not accurate.
The quality evaluation of such images has been proved challenging for IQA methods \cite{pipal}.
Gu \etal \cite{pipal} contribute an IQA dataset called Perceptual Image Processing ALgorithms dataset (PIPAL), including the results of Perceptual-oriented image processing algorithms.
This data set is used to benchmark different IQA methods and is used as the training and testing dataset in this challenge.

\begin{figure*}
    \centering
    \includegraphics[width=\linewidth]{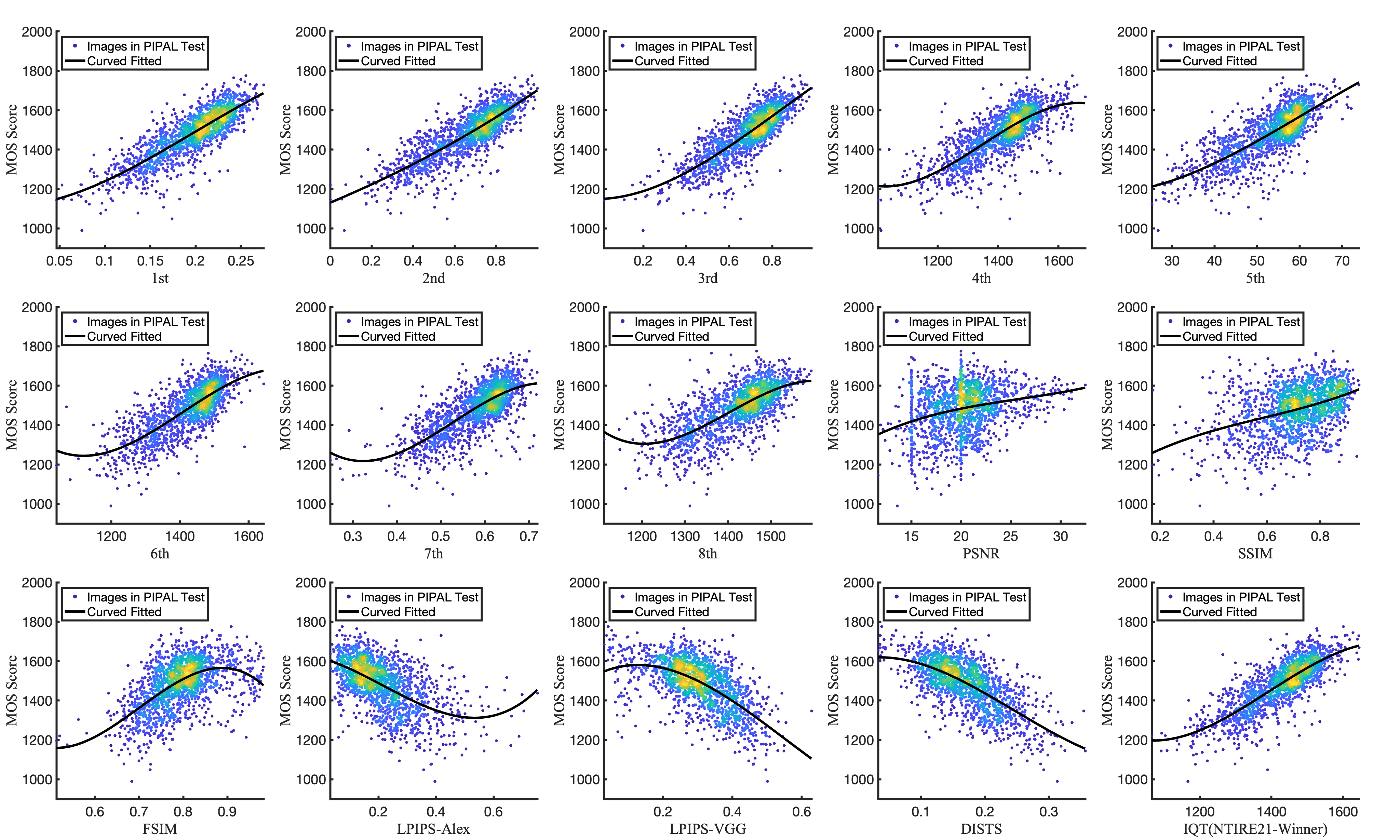}
    \caption{FR-IQA Track's Scatter plots of the objective scores vs. the MOS scores. The curves were obtained by a third-order polynomial nonlinear fitting.}
    \label{fig:4.1.1}
\end{figure*}

\begin{figure*}
    \centering
    \includegraphics[width=\linewidth]{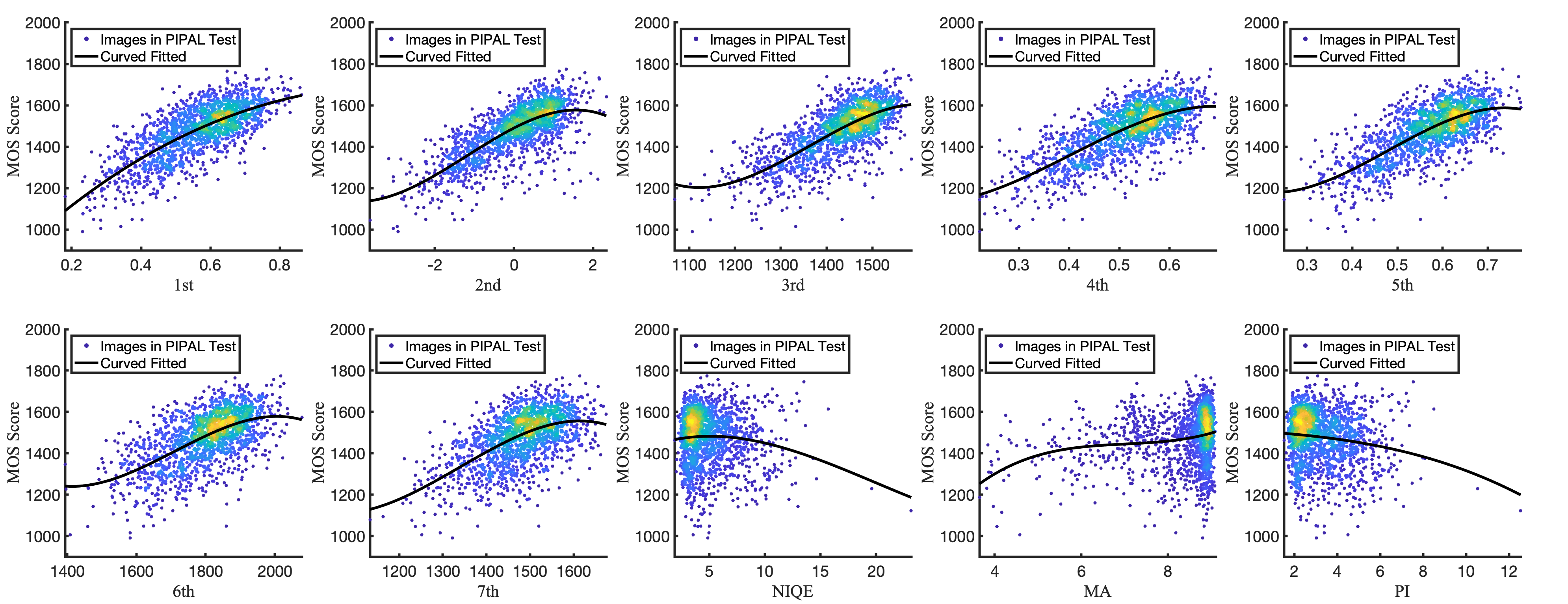}
    \caption{NR-IQA Track's Scatter plots of the objective scores vs. the MOS scores. The curves were obtained by a third-order polynomial nonlinear fitting.}
    \label{fig:4.1.2}
\end{figure*}

\begin{figure*}
    \centering
    \includegraphics[width=\linewidth]{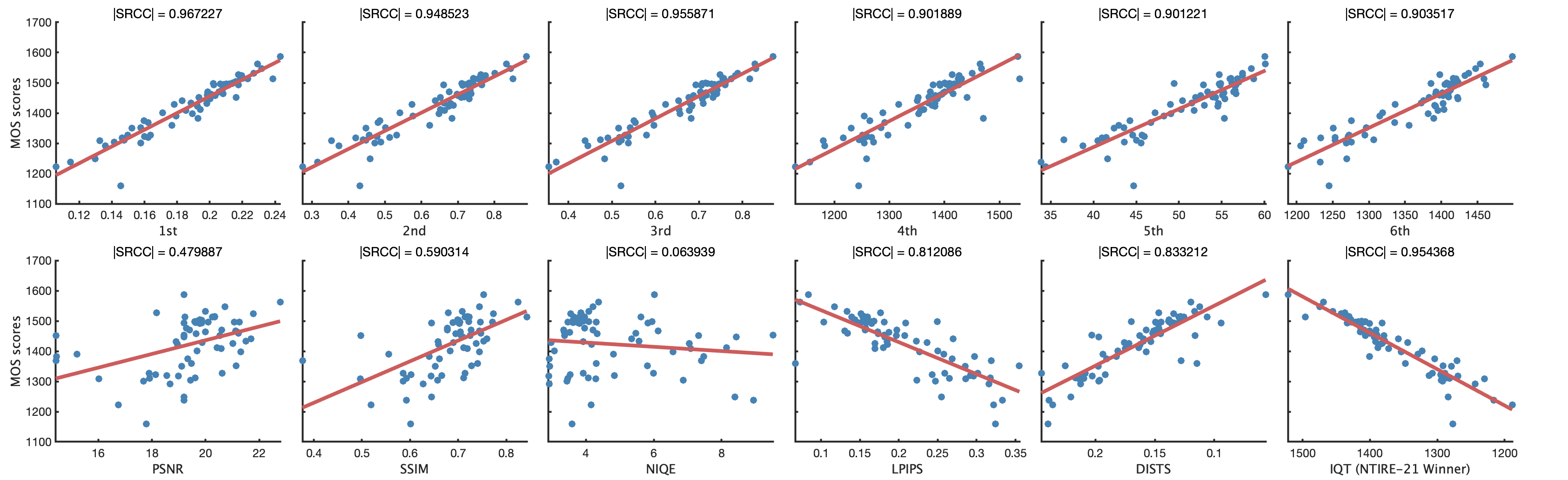}
    \caption{Analysis of FR-IQA methods in evaluating IR methods. Each point represents an algorithm. Higher correlations indicates better performance in evaluating perceptual image algorithm.}
    \label{fig:4.1.3}
\end{figure*}

\begin{figure*}
    \centering
    \includegraphics[width=\linewidth]{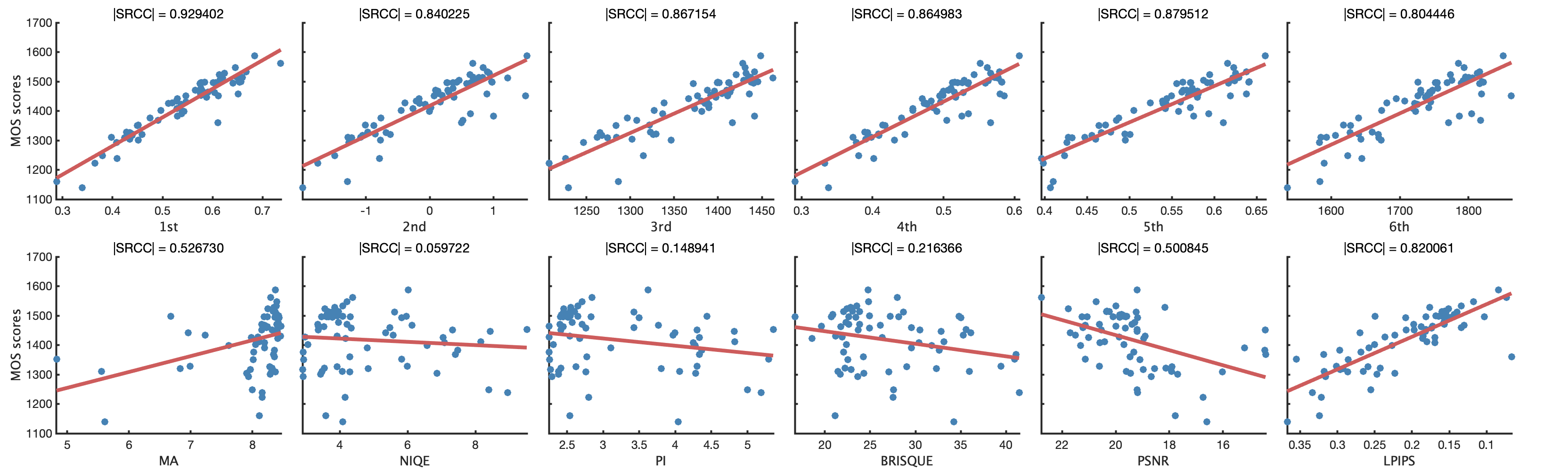}
    \caption{Analysis of NR-IQA methods in evaluating IR methods. Each point represents an algorithm. Higher correlations indicates better performance in evaluating perceptual image algorithm.}
    \label{fig:4.1.4}
\end{figure*}

%-------------------------------------------------------------------------
\section{The NTIRE Challenge on Perceptual IQA}
We host the NTIRE 2022 Perceptual Image Quality Assessment Challenge to push developing state-of-the-art FR- and NR- IQA methods to deal with the novel GAN-based distortion types, compare different solutions, and gain new insights.
Details about the challenge are as follows:

\paragraph{Tracks.}
We include two tracks: the FR-IQA track that focuses on evaluating full-reference IQA methods and a new NR-IQA track for no-reference IQA methods.

\begin{itemize}
    \item Track 1: The task of this track is to obtain an FR-IQA method capable of producing high-quality perceptual similarity results between the given distorted images and the corresponding reference images with the best correlation to the reference ground truth MOS score.
    \item Track 2: The task of this track is to obtain an NR-IQA method capable of producing high-quality perceptual quality results with the best correlation to the reference ground truth MOS score. Only distorted images are given in this track, and no reference images are available. 
\end{itemize}

\paragraph{Dataset.}
Following NTIRE 2021 IQA challenge \cite{gu2021ntire}, we employ a subset of the PIPAL dataset as the training set and an extended version of the PIPAL dataset as the validation and the testing set.
The PIPAL dataset includes traditional distortion types, image restoration results, compression results, and novel GAN-based image processing outputs.
More than 1.88 million human judgements are collected to assign mean opinion scores (MOS) for PIPAL images using the Elo rating system \cite{elo1978rating}.
The original PIPAL dataset includes 250 high-quality, diverse reference images, and each has 116 different distorted images.
We use 200 of the 250 reference images and their distorted images as the training set (in total $200\times116$ distorted images).
All training images and the MOS scores are publicly available.

The validation set and the testing set are selected from the extended version of the PIPAL dataset \cite{gu2021ntire}.
The validation set contains 25 reference images and 40 distorted images for each.
The testing set contains 25 reference images and all the 66 distorted images for each reference image.
The newly collected distortion types are all outputs of GAN-based image restoration algorithms or GAN-based compression algorithms. In total, 3300 additional images are collected.
Note that for the participants, the training set and the validation/testing set contain completely different references and distorted images, ensuring the final results' objectivity.
Methods trained with additional labelled IQA datasets (pre-training using non-IQA datasets such as ImageNet is allowed) will be disqualified from the final ranking for both tracks.

\paragraph{Evaluation protocol.}
Align with the challenge at NTIRE 2021 \cite{gu2021ntire}, our evaluation indicator, namely main score, consists of both Spearman rank-order correlation coefficient (SRCC) \cite{sheikh2006statistical} and Person linear correlation coefficient (PLCC) \cite{benesty2009pearson}:
\begin{equation}
	\mathrm{Main\ Score}=|\mathrm{SRCC}|+|\mathrm{PLCC}|.
\end{equation}
The SRCC evaluates the monotonicity of methods to find whether the scores of high-quality images are higher (or lower) than low-quality images.
The PLCC is often used to evaluate the accuracy of methods \cite{sheikh2006statistical,gu2020image}.
Before calculating PLCC index, we perform the third-order polynomial nonlinear regression as suggested in the previous works \cite{tid2013,pipal}.
By combining SRCC and PLCC, our indicator can measure the performance of participating models in an all-around way.

\paragraph{Challenge phases.}
The whole challenge consists of three phases: the developing phase, the validation phase, and the testing phase.
In the developing phase, the participants can access to the reference and distorted images of the training set and also the MOS labels.
This period is for the participants to familiarize themselves with the data structure and develop algorithms.
In the validation phase, the participants can access the reference and distorted images of the training set and no labels are provided.
The participants had the opportunity to test their solutions on the validation images and to receive immediate feedback by uploading their results to the server.
A validation leaderboard is available.
In the testing phase, the participants can access to the reference and distorted images of the training set.
A final predicted perceptual similarity result is required before the challenge deadline.
The participants also need to submit the executable file and a detailed description file of the proposed method. 
The final results were then made available to the participants.

%-------------------------------------------------------------------------
\section{Challenge Results}
\label{sec:results}
There are 8 and 7 teams participated in the testing phase of the challenge for the track 1 and track 2, respectively.
\tablename~\ref{tab:main_results} reports the main results and important information of these teams.
The methods are briefly described in Section~\ref{sec:methods} and the team members are listed in Appendix~\ref{sec:apd:track1team} and Appendix~\ref{sec:apd:track2team}.
We next analyze each track's result separately

\subsection{Track 1: Full-Reference IQA Track}
This is the second full-reference IQA challenge.
In the last challenge, IQT \cite{IQT2021ntire} won the championship on this track using a transformer as the network backbone.
This year, we use a more complex validation and testing dataset.
According to the results in \tablename~\ref{tab:main_results}, we can see that this year's submitted methods have generally achieved comparable results.
All valid entries achieved higher correlation performance than methods such as LPIPS \cite{zhang2018unreasonable}, which are now widely used.
Three teams surpassed last year's championship method.
The champion team achieves an SRCC score of 0.823 and a PLCC score of 0.828, refreshing the state-of-the-art performance on PIPAL.
\figurename~\ref{fig:4.1.1} shows the scatter distributions of subjective MOS scores vs. the predicted scores by the top solutions and the other 7 IQA metrics on the PIPAL test set.
The curves shown in \figurename~\ref{fig:4.1.1} were obtained by a third-order polynomial nonlinear fitting.
One can observe that the objective scores predicted by the top solutions have higher correlations with the subjective evaluations than existing methods.
In \figurename~\ref{fig:4.1.3}, we show the scatter plots of subjective scores vs. the top solutions and some commonly-used IQA metrics for perceptual-oriented algorithms.
Recall that an important goal of this challenge is to promote more promising IQA metrics for perceptual-oriented algorithms.
As can be seen, the top solutions generally perform better in evaluating the images in the testing set.
Among them, the correlation between the evaluation of the champion's solution and the subjective score reaches 0.967, which surpasses the champion's performance of the last year.

\subsection{Track 2: No-Reference IQA Track}
It is the first time an NTIRE challenge focuses on no-reference IQA.
NR-IQA is an indispensable part of algorithm evaluation, but widely used algorithms only show very limited performance on our test set.
This year, we include this track to push the developing state-of-the-art NR-IQA methods to fill this gap.
According to the results in \tablename~\ref{tab:main_results}, one can observe that all the valid entries surpass the current state-of-the-art performance, and some even achieve correlation performance comparable to FR methods.
\figurename~\ref{fig:4.1.2} shows the scatter distributions of subjective MOS scores vs. the predicted scores by the top solutions and the other 3 NR-IQA metrics.
The curves show compatible conclusions.
In \figurename~\ref{fig:4.1.4}, we show the scatter plots of subjective scores vs. the top solutions and some commonly-used IQA metrics for perceptual-oriented algorithms.
It can be seen that the existing NR-IQA methods are not ideal in evaluating algorithms.
The works produced in this challenge received high correlation scores, which means that these methods are closer to human judgment when used to evaluate images generated by perceptual-oriented algorithms.
This also suggests that using these NR methods as metrics can lead to more visually friendly results.
Among them, the correlation between the evaluation of the champion solution and the subjective score reaches 0.92, which greatly improves the practical value of NR-IQA as an algorithm metric.

\begin{figure*}
    \centering
    \includegraphics[width=\linewidth]{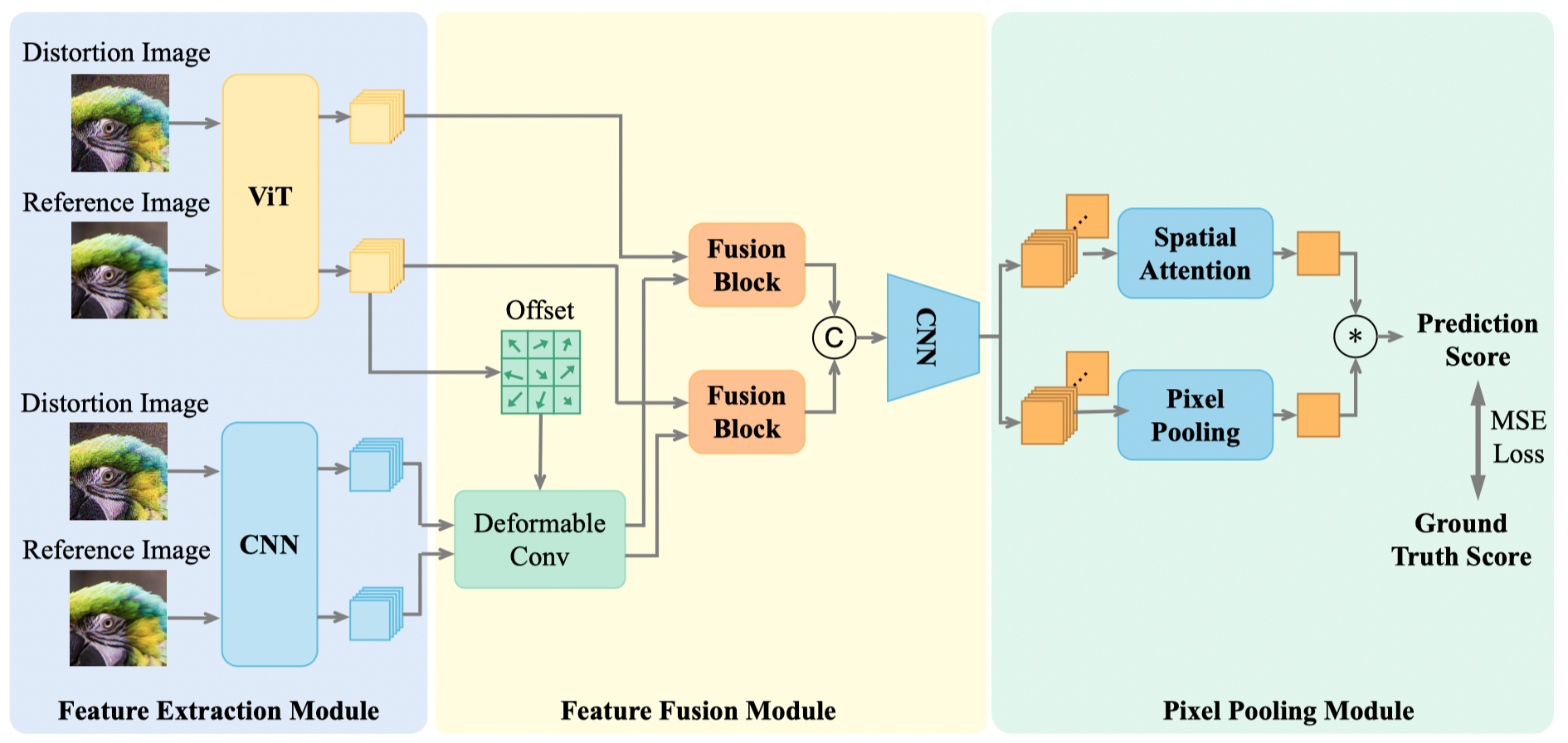}
    \caption{The overview of THUIIGROUP1919 team's Attention-based Hybrid Image Quality assessment network (AHIQ).}
    \label{fig:5.1.1}
\end{figure*}

\begin{figure}
    \centering
    \includegraphics[width=\linewidth]{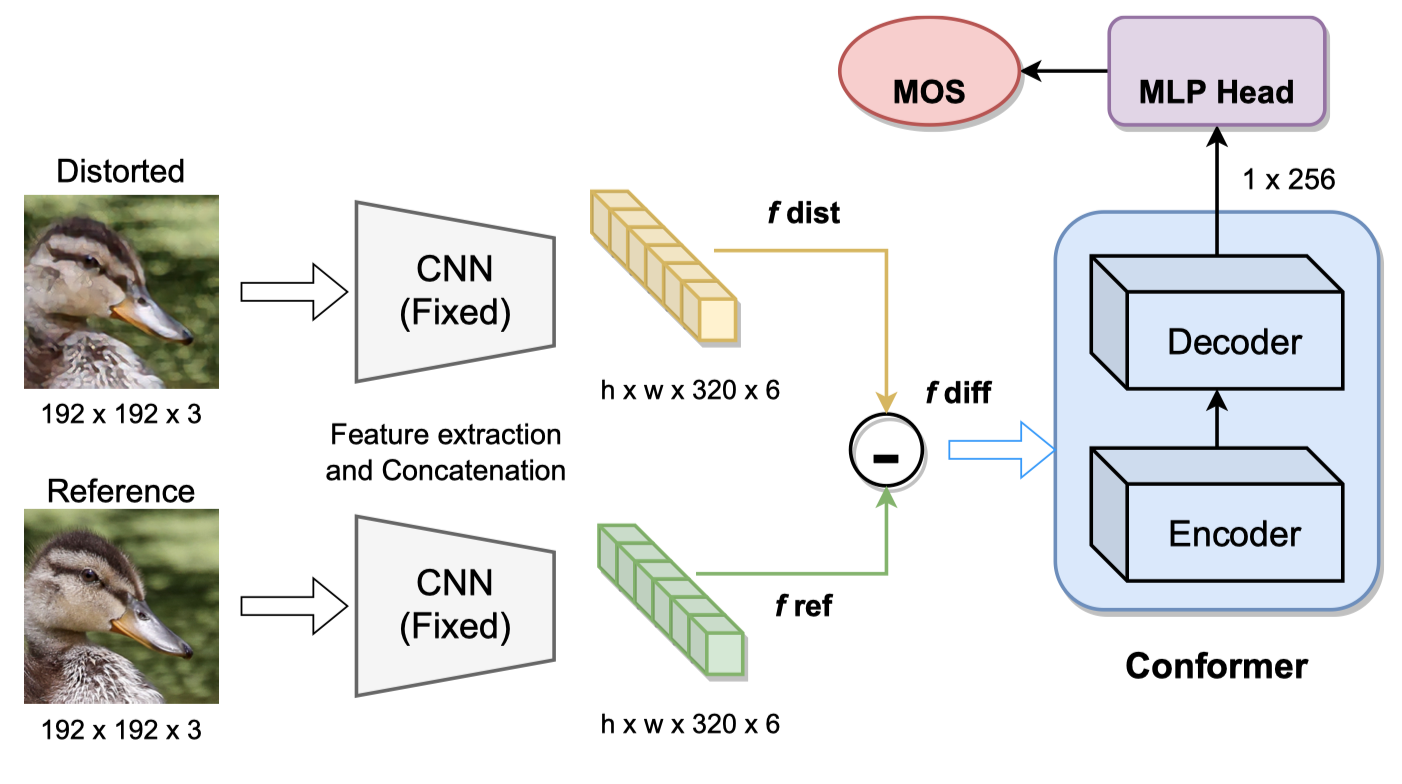}
    \caption{The network design of JMU-CVLab team's method.}
    \label{fig:5.1.4}
\end{figure}

\begin{figure*}
    \centering
    \includegraphics[width=\linewidth]{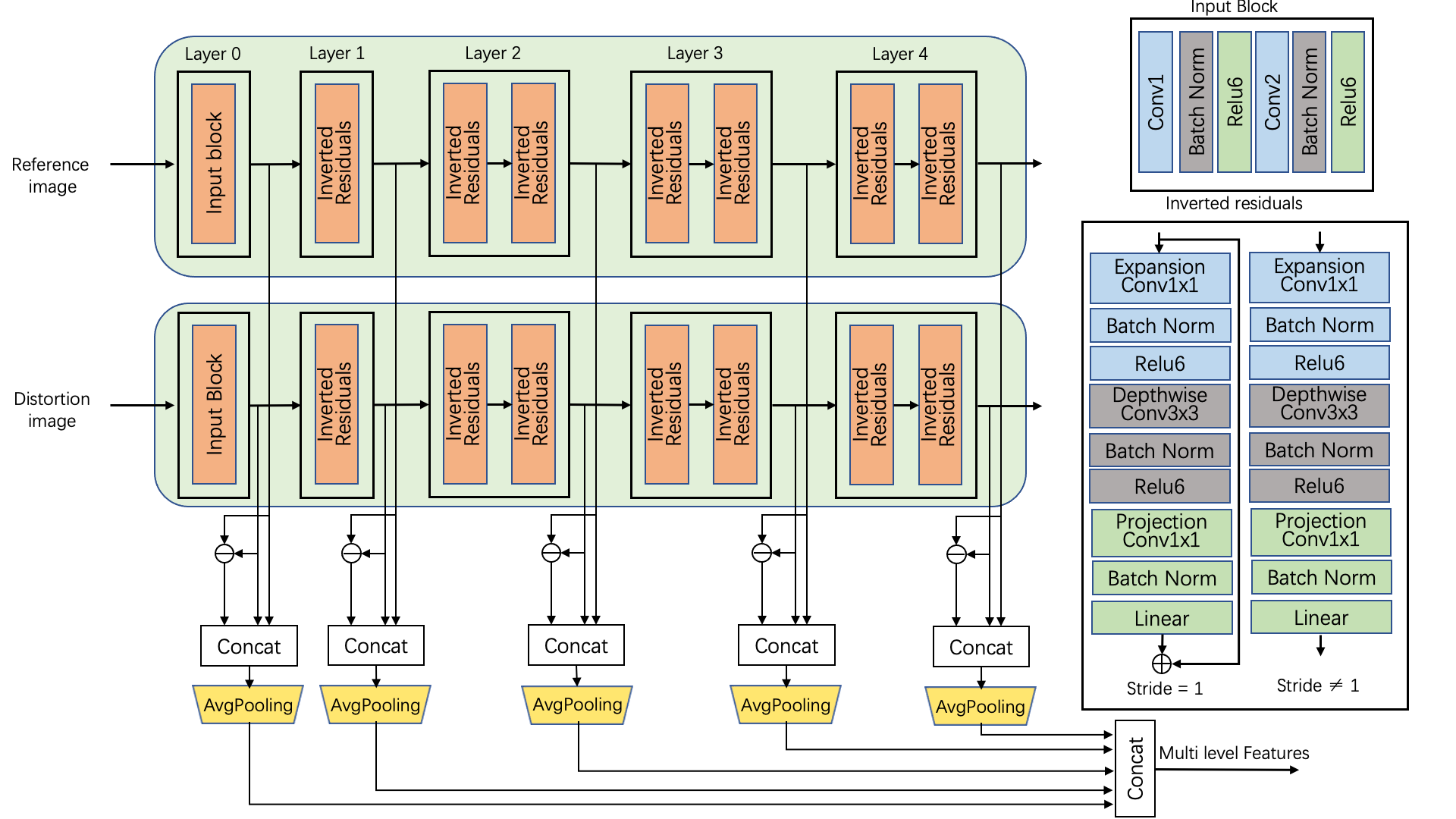}
    \caption{The framework design for feature extraction of Yahaha! team's solution.}
    \label{fig:5.1.5(1)}
\end{figure*}

\begin{figure*}
    \centering
    \includegraphics[width=\linewidth]{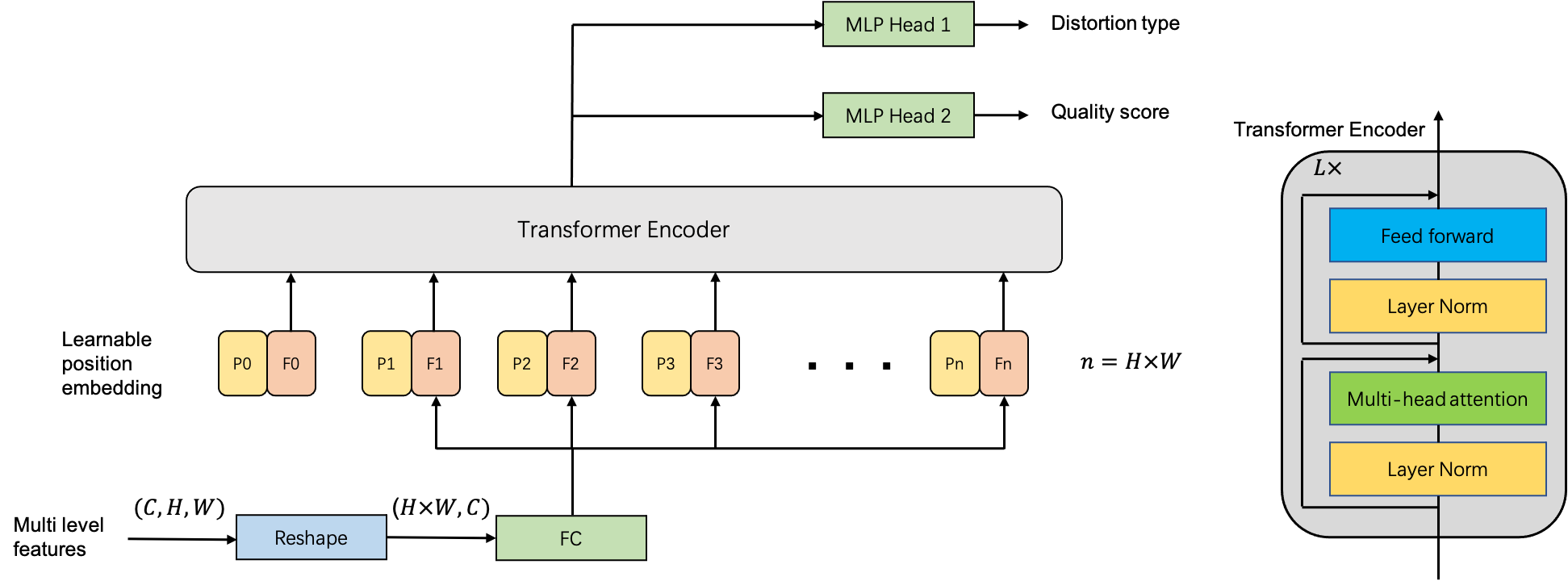}
    \caption{The framework of transformer of Yahaha! team's solution.}
    \label{fig:5.1.5(2)}
\end{figure*}

\begin{figure}
    \centering
    \includegraphics[width=\linewidth]{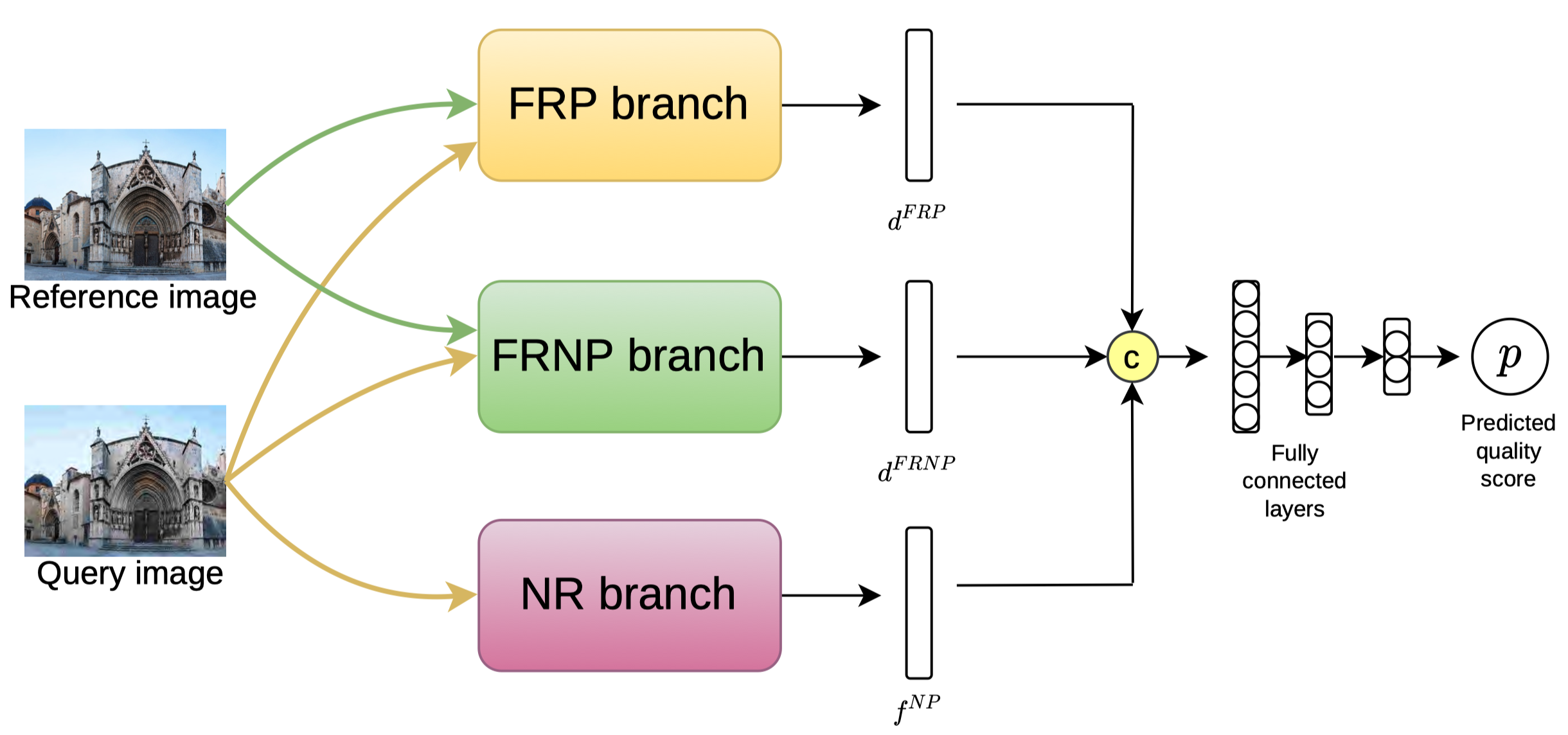}
    \caption{The framework design of the Horizon team's method.}
    \label{fig:5.1.7}
\end{figure}

\begin{figure*}
    \centering
    \includegraphics[width=\linewidth]{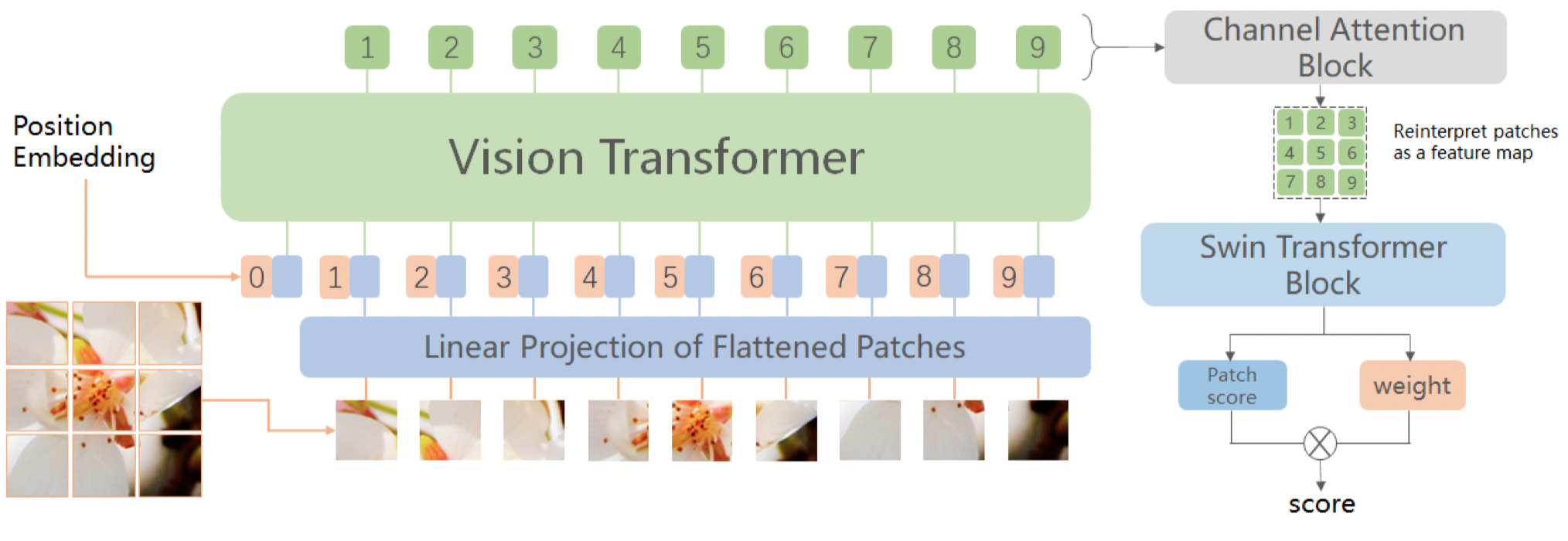}
    \caption{The model framework overview of the THU\_IIGROUP team.}
    \label{fig:5.2.1}
\end{figure*}

\begin{figure}
    \centering
    \includegraphics[width=\linewidth]{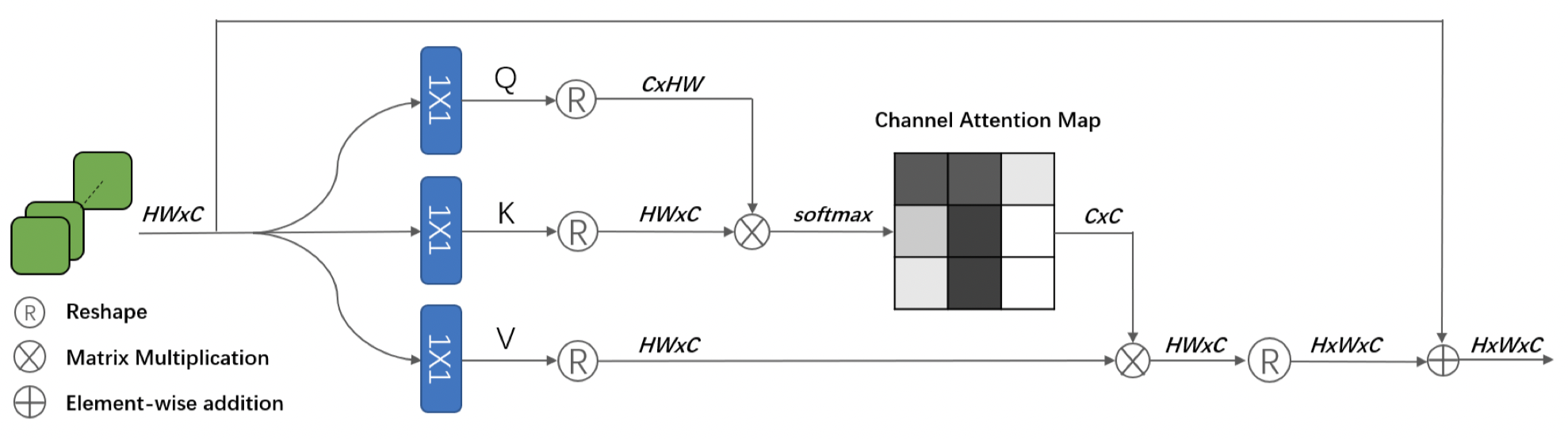}
    \caption{The Channel attention block used in THU\_IIGROUP team's network.}
    \label{fig:5.2.1(1)}
\end{figure}

\begin{figure}
    \centering
    \includegraphics[width=\linewidth]{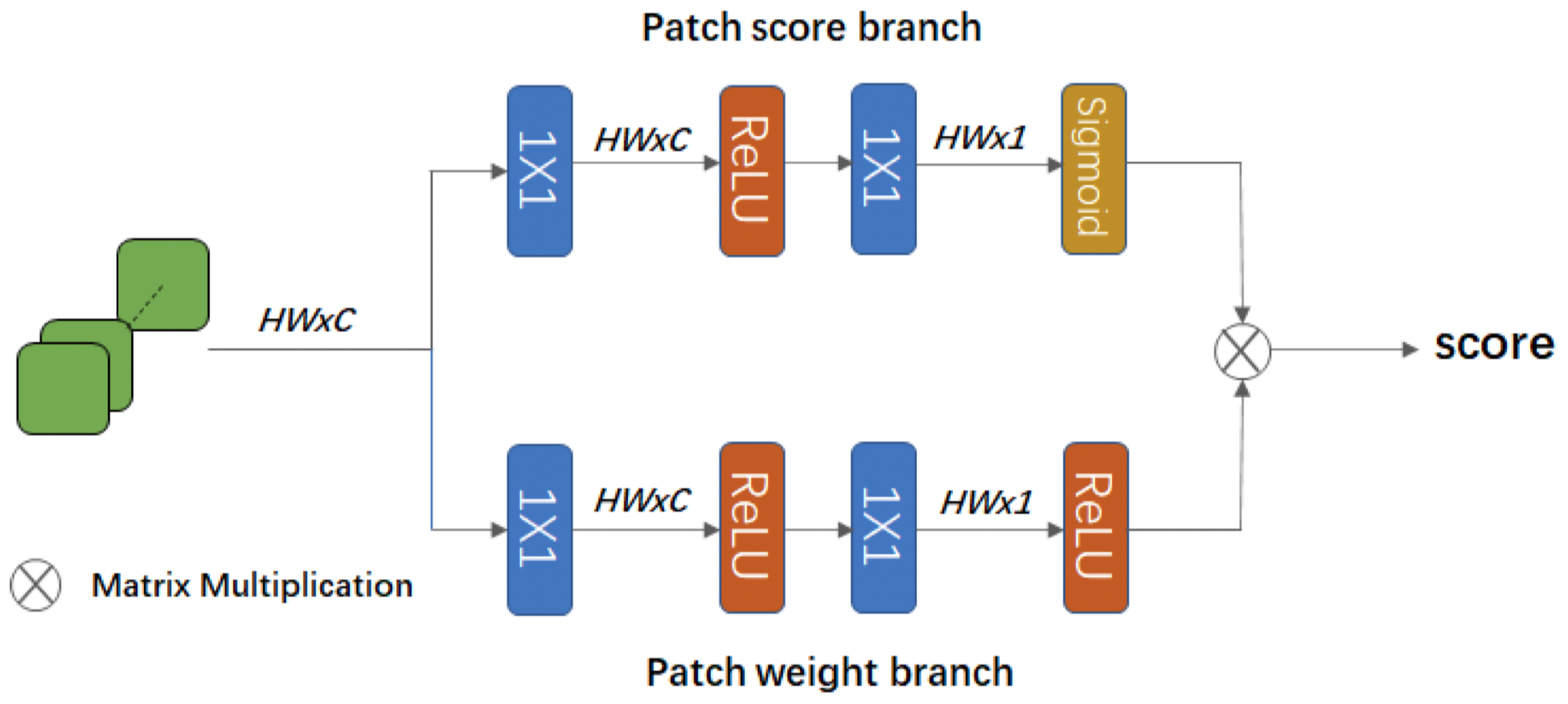}
    \caption{The Patch Weighted Branch used in THU\_IIGROUP team's network.}
    \label{fig:5.2.1(2)}
\end{figure}

\begin{figure}
    \centering
    \includegraphics[width=\linewidth]{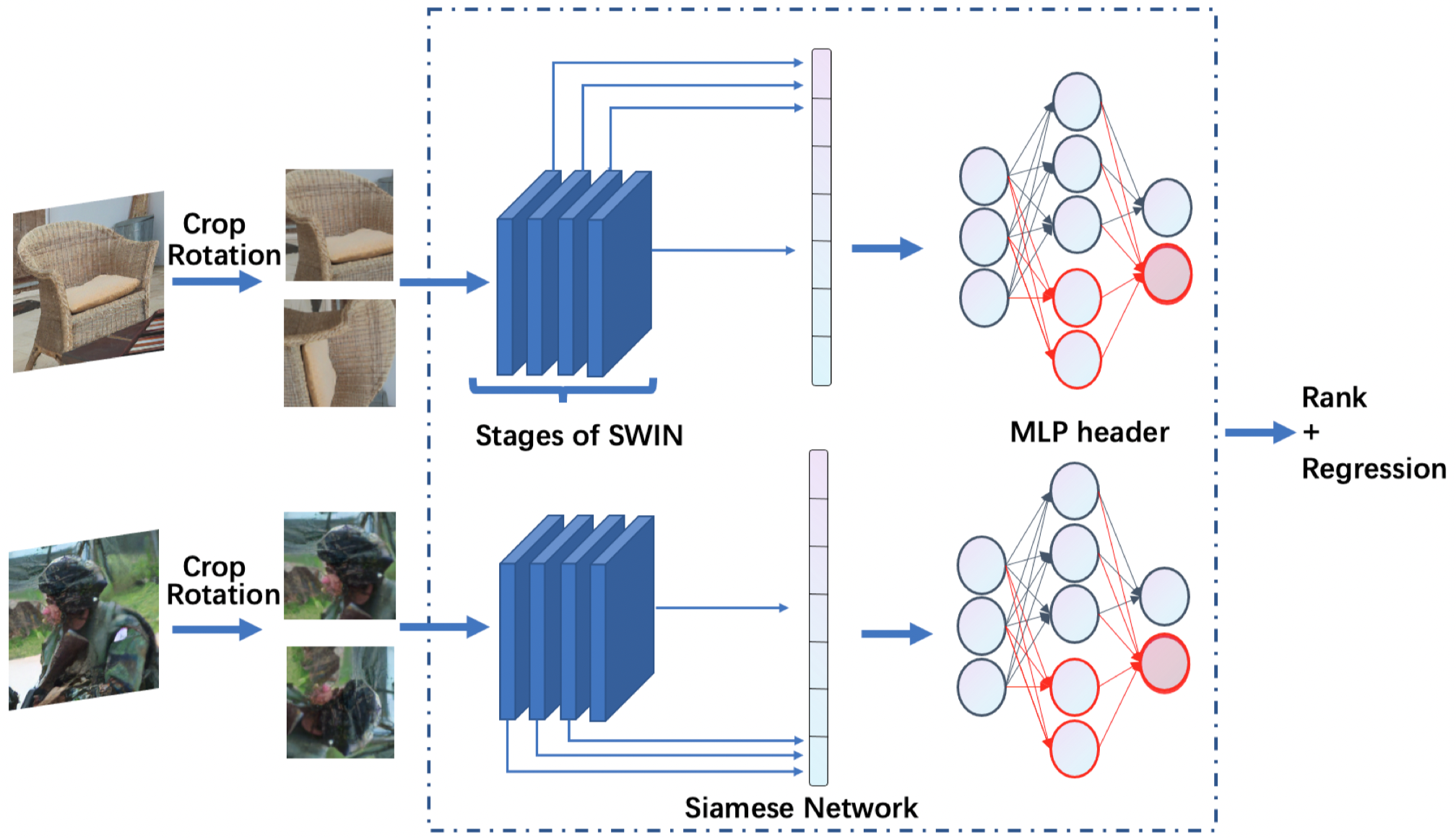}
    \caption{The Multi Stage fusing SWIN Transformer based on Siamese Network design proposed by team DTIQA.}
    \label{fig:5.2.2}
\end{figure}

\begin{figure}
    \centering
    \includegraphics[width=\linewidth]{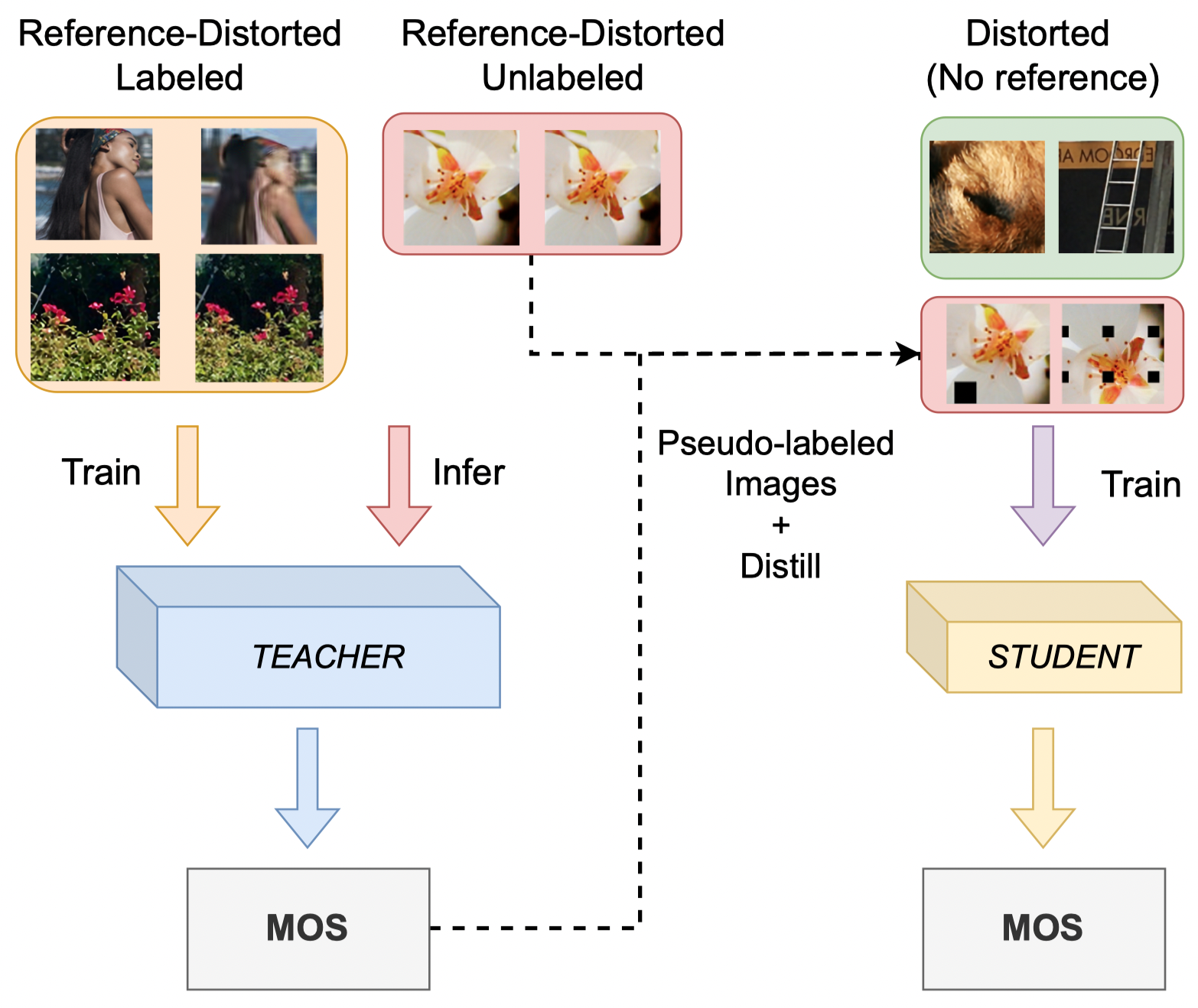}
    \caption{The blind noisy student setup proposed by team JMU-CVLab.}
    \label{fig:5.2.3}
\end{figure}

\begin{figure}
    \centering
    \includegraphics[width=\linewidth]{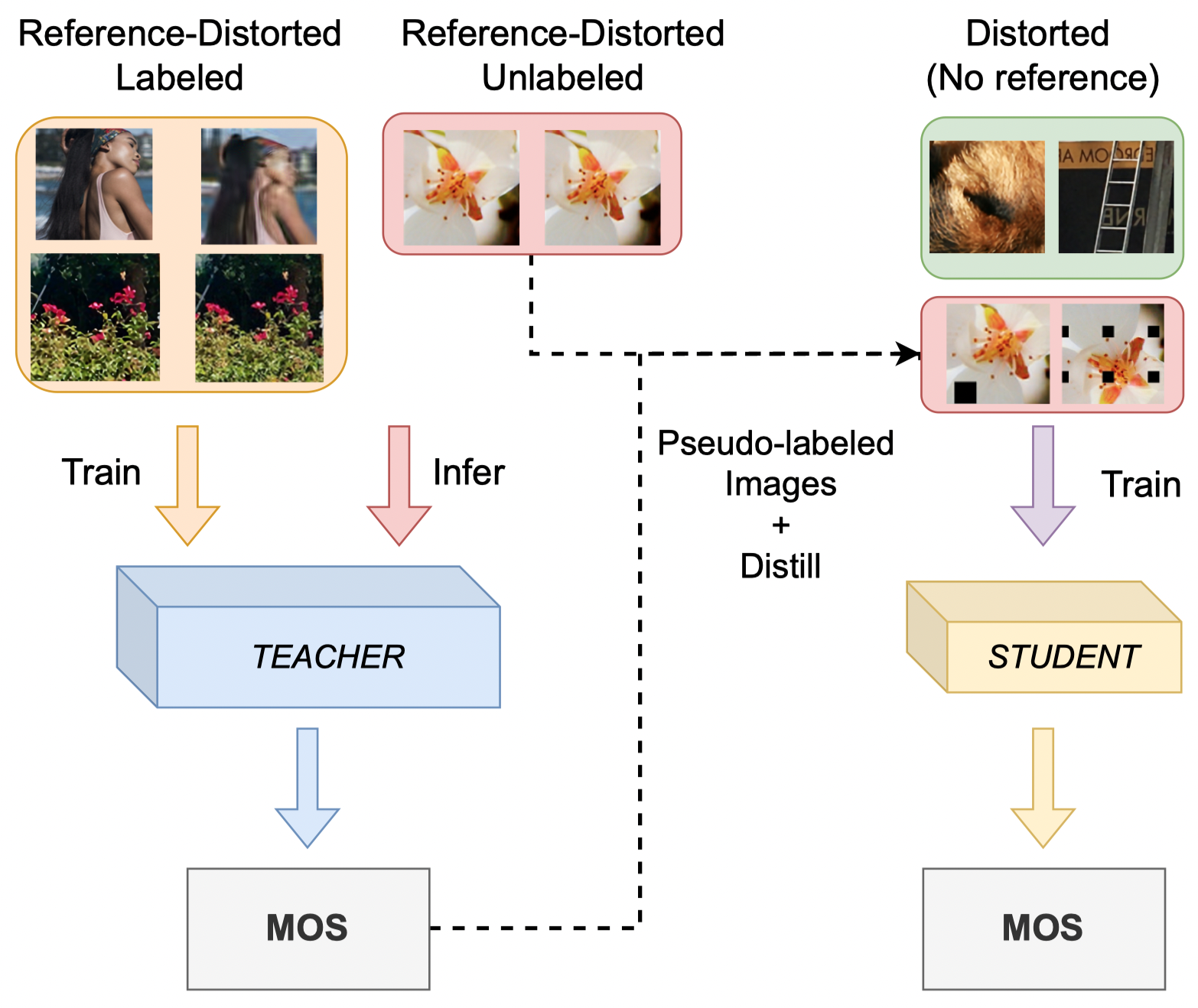}
    \caption{The blind noisy student setup proposed by team JMU-CVLab.}
    \label{fig:5.2.3}
\end{figure}

\begin{figure}
    \centering
    \includegraphics[width=\linewidth]{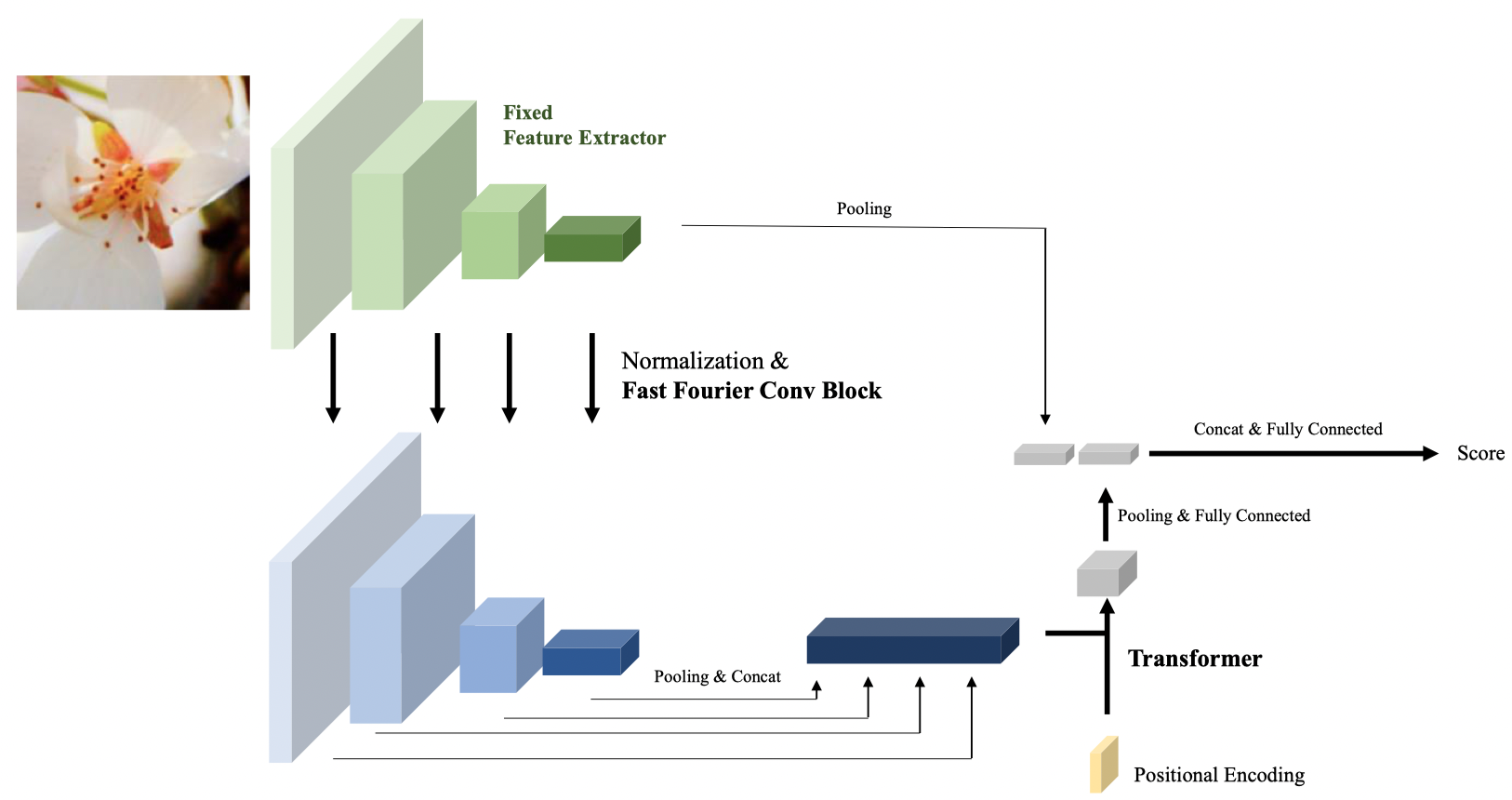}
    \caption{The method of an anonymous team.}
    \label{fig:5.2.5}
\end{figure}

%------------------------------------------------------------------------
\section{Challenge Methods}
\label{sec:methods}
We describe the submitted solution details in this section.

\subsection{Track 1: Full-Reference IQA Track}

\subsubsection{THU1919Group}
Team THU1919Group is the winner of the first track.
They develop an Attention-based Hybrid Image Quality assessment network (AHIQ) to participate in the FR-IQA track.
They also submit a challenge paper for their method \cite{2022Attentions}.
As is shown in \figurename~\ref{fig:5.1.1}, their network takes pairs of reference images and distortion images as input and consists of three key components: the feature extraction module, the feature fusion module and the pixel pooling module.
For the feature extraction module, the input pairs of reference images and distortion images first go through a vision transformer backbone ViT \cite{dosovitskiy2020image} and a convolution network (CNN) \cite{he2016deep} for feature extraction.
The convolution network retains more spatial information, and the transformer network captures global semantic features.
In the feature fusion module, the feature maps from later stages of ViT are used to obtain an offset map for deformable convolution.
A simple 2-layer convolution network is used to project the feature after deformable convolution.
In this way, features from the early stages of CNN can be better modified and utilized for further feature fusion. 
At last, they propose the pixel pooling module to assess the quality of distorted images.
The pixel pooling module contains two branches.
The first branch calculates scores for each pixel, and the second branch calculates the weight of each pixel score to the final evaluation score.
By weighting all the pixel scores, the final score can be obtained.

Their network contains 140 million parameters.
For optimization, they use the AdamW \cite{loshchilov2017decoupled} optimizer with an initial learning rate(LR) of $10^{-4}$ and weight decay of$10^{-5}$.
The minibatch size is 8.
During testing, different backbone networks and fusion approaches are used for ensemble, such as models at different training epochs, directly concatenate the feature maps from CNN and ViT without deformable convolution, and use inception-resnetV2 as feature extraction and use ResNet152 as feature extraction.

\subsubsection{Netease OPDAI}
Team Netease OPDAI wins second place in the first track.
They also submit a challenge paper for their method \cite{2022Image}.
They build their method based on the winner solution of the last year -- the IQT network \cite{IQT2021ntire}.
The difference is that they concatenate the reference image and the distorted image instead of the difference operation.
They also introduce central difference convolution and Siamese network structure to extract features of the distorted image and reference image, respectively.
They use Swin \cite{liu2021swin} transformer as regression layer.
Finally, they incorporate a residual network using the spatial gradient module.

For optimization, in addition to the conventional MSE loss, they also learn the distribution of quality scores by introducing the Kullback–Leibler scatter loss, and norm-in-norm loss \cite{li2021reproducibility}.
They also introduce three data enhancement methods to help training.
First, they dynamically optimize the frequency of difficult samples depending on the model ﬁtting situation.
Second, the model must adapt to random color space variations to improve generalization.
And third, different positions of the distorted images are stitched together to increase the complexity of the dataset and to improve the robustness of the model.
Their method has more than 276 million parameters.

\subsubsection{KS}
Team KS wins third place in the first track.
They apply a bag of tricks for the IQA method with newly-appeared distortions.
Firstly, they designed a new Multi-Scale Image Quality Network, called MSIQ-Net, to fully capture spatial distributions of distortion characteristics.
MSIQ-Net takes a pair of distortion and reference images as input and generates multi-scale features using an FPN-like module.
Among different scales, local texture distortion (\eg, noise and blocking artifact) can be captured by low-level features, while global distributed distortion (\eg, strange artifacts generated by GAN) can be learned from high-level features.
A smooth module aggregating features adaptively is also attached for the final representation.

They attach great importance to data processing during training.
They discover that the labels of the training set have an unbalanced distribution.
To prevent the model from being biased, they reconstruct the training data.
First, for images with low/high MOS scores, they perform data augmentations (\eg, horizontal flip, rotation) and increase the number of these images in the training set.
Second, they follow the way in PIPAL and select high-frequency patches from SPAQ \cite{fang2020perceptual}.
They then generate pseudo labels for these patches using a model trained on the given data.
These patches, along with pseudo-labels, are added to the training process.
Third, they apply a histogram equalization on labels and obtain an even distribution.

For the loss functions, in addition to the commonly used MSE loss, they also employ an extra PLCC-induced loss \cite{li2021reproducibility}.
Assume we have $N$ images in the training batch.
Given the predicted quality scores $Y' = \{ y_1',\dots, y_N'\}$ and the subjective quality scores $Y = \{y_1,\dots,y_N \}$ , the loss is defined as
$$
\mathcal{L}_{plcc}=\big(1-\frac{\sum_{i=1}^N (y'_i-a')(y_i-a)}{\sqrt{\sum_{i=1}^N (y'_i-a')^2\sum_{i=1}^N (y_i-a)^2}}\big)*0.5,
$$
where $a'$ and $a$ are the mean values of $Y'$ and $Y$, respectively.

The KS team also adopt the model ensemble strategy.
Three main methods are used.
First, by directly averaging the weights of multiple models trained with different hyper-parameters, the IQA model improves accuracy and robustness.
Second, due to the model capacity and divergence, a single model may not make the perfect predictions for a given dataset, suffering from specific noise or bias. The combination can be implemented by averaging the output of each model. A weighted combination will also do the job, whose weights can come from linear regression or other schemes.
Third, various augmentation types, which do not inflect the quality of images, are used for repeat predictions, including multi-crop, horizontal flip and, random rotation. The average score of multiple views is used for the final prediction.

\subsubsection{JMU-CVLab}
Team JMU-CVLab~\cite{conde2022conformer} proposes an IQA Conformer Network \cite{2022Conformer} by improving the IQT \cite{IQT2021ntire} architecture.
They use Inception-ResNet-v2 \cite{szegedy2017inception} network pre-trained on ImageNet \cite{deng2009imagenet} to extract reference and distorted images feature maps.
The network weights are kept frozen, and a Conformer encoder-decoder is trained to regress MOS using the MSE loss.
In their network, the \texttt{mixed5b}, \texttt{block35\_2}, \texttt{block35\_4}, \texttt{block35\_6}, \texttt{block35\_8} and \texttt{block35\_10} feature maps are concatenated for the reference and distorted images generating $f_{ref}$ and $f_{dist}$, respectively.
In order to obtain the difference information between reference and distorted images, a difference feature map, $f_{diff}=f_{ref}-f_{dist}$ is also used.
Concatenated feature maps are then projected using a point-wise convolution but not flattened to preserve spatial information.
They used a single Conformer block \cite{burchi2021efficient,gulati2020conformer} for both encoder and decoder. 
The model hyper-parameters are: $L = 1$, $D = 128$, $H = 4$, $D_{feat} = 512$, and $D_{head}= 128$.
The input image size of the backbone model is set to $192\times192\times3$, which generates feature maps of size $21\times21$.
Their network design is shown in \figurename~\ref{fig:5.1.4}.
They also adopt the ensemble method to improve the performance of the final method.
Their final submission is an ensemble of the proposed IQA Conformer Network and two pre-trained models: RADN~\cite{RADN2021ntire}, and ASNA~\cite{ayyoubzadeh2021asna}.

\subsubsection{Yahaha!}
Team Yahaha! proposes a transformer-based full-reference image quality assessment framework leveraging multi-level features.
The reference image and distortion image are fed to the backbone network separately. Their feature maps from different layers and difference maps of corresponding feature maps are downsampled and concatenated together.
Then all these maps from different levels are fed to transformer layers for the joint training of distortion type prediction and perceptual score regression.
A Siamese-network is used for extracting deep features from the reference image and distorted image, as shown in \figurename~\ref{fig:5.1.5(1)}.
A pruned MobileNetv2 is used in this stage, and feature maps from 4 layers of MobileNetv2 are extracted for further processing.
Then the extracted features are fed to a Transformer network to predict the final score, the architecture of which is shown in \figurename~\ref{fig:5.1.5(2)}.
The Transformer encoder contains multi Transform layers, each consisting of a standard multi-head attention module and a feedforward module. Besides, layer normalization is adopted. The Transformer encoder is connected with two different fully connected layers to predict distortion type and opinion score, respectively.

For the optimization of the proposed method, they use a loss function consisting of four loss functions: L1 loss, cross-entropy loss, norm-in-norm loss \cite{li2020norm} and relative-distance loss.
The number of parameters for their proposed model is 68.853K.
They also adapt ensemble strategy.
The entire training set is divided into five parts. These five parts are used as the validation set, while the other four parts are used for training.
After obtaining five models, the submitted prediction scores for development and test are the average prediction results of these five models.

\subsubsection{debut\_kele}

The debut\_kele team's solution can be divided into two main parts: feature extraction and regression modeling.
For the features, they extract different perceptual image quality metrics, which include SSIM, MS SSIM, CW-SSIM, GMSD, LPIPSvgg, DISTS, NLPD, FSIM, VSI, VIFs, VIF, MAD.
Using gradient boosting trees, a regression model is built based on the pre-calculated IQA results.
The model's `max depth' parameter is set to 3, and the learning rate was set to 0.01.
In addition, the feature subsample and the sample subsample values were set at 0.7 to prevent overfitting.
The maximum iteration round is 10000, while the early-stopping round is 500.
In their experiments, VIFs and VIF play a more critical role than LPIPSvgg.
The five-fold bagging-based ensemble strategy is used in their experiments.

\subsubsection{Pico Zen}
The Pico Zen team proposes Multi Scale Image Quality Assessment with Transformers by weight sharing approach \cite{2022multi}.
They take the baseline model from \cite{IQT2021ntire} and introduce a weight-sharing method.
Using the above approach, they take four different scales of images i.e. downscale image by factor 2 (Scale2), downscale image by factor 3 (Scale3), upscale image by factor 2 (Scale0.5), and original image for the ensemble.
Their method is described in \cite{2022multi}.

\subsubsection{Team Horizon}
The Horizon team proposes a Multi-branch Image Quality Assessment Network \cite{2022can} that consists of three parallel branches: (1) the full-reference pre-trained (FRP) branch, (2) the full-reference non-pretrained (FRNP) branch and (3) the no-reference (NR) branch.
Both distorted and reference images are fed as input for the full-reference branches (FRP and FRNP), whereas in the no-reference branch, only the distorted image is provided as input.
In each branch, they use a convolutional neural network-based encoder.
In the FRP branch, they use an encoder from a classifier trained on ImageNet since these features are known to correlate well with perceptual quality.
The weights of the FRP encoder are kept fixed throughout the training.
They didn't use pre-trained encoders in FRNP and NR branches to enable the learning of discriminative features from the training data.
Full-reference branches extract features from both distorted and reference images and compute pixel-wise differences between distorted and reference features.
But when the distortion is severe, computing pixel-wise difference, even in feature space, may not be optimal due to spatial misalignment.
Hence, a No-reference branch focuses only on features related to the distortion present in the query image.
The framework of the proposed method is shown in \figurename~\ref{fig:5.1.7}.

\subsection{Track 2: No-Reference IQA Track}

\subsubsection{THU\_IIGROUP}
Team THU\_IIGROUP is the winner of the second track.
They also submit a challenge paper for their method \cite{2022MANIQA}.
In their method, they first cut of the image with a small size generates the global interaction between different regions of an image. It utilizes the pre-trained vision transformer \cite{dosovitskiy2020image} as the feature extractor to attain the feature of different patches from different positions.
They select four layers of the vision transformer output as the main features.
Second, to handle the difference among four layers, a channel attention block \cite{zamir2021restormer} is used to adjust the distribution of features from these four layers.
Then, to reinforce the local connection with image patches, a Swin transformer block \cite{liu2021swin} is applied to handle the perceptual quality of each patch.
Finally, a regression head with two branches -- patch score branch and weight branch \cite{bosse2017deep} -- is used to predict the quality score and the importance of each patch. A multiplication generates the final score of the distorted image.
The overview of the method is shown in \figurename~\ref{fig:5.2.1}, the channel attention block is shown in \figurename~\ref{fig:5.2.1(1)}, and the final patch weighted branch is shown in \figurename~\ref{fig:5.2.1(2)}.
They also employ a bagging ensemble method.
Three same models are trained for ensembles. (1) The model M1 without finetuning. (2) The model M2 finetune twice. (3) The model M3 finetune only once. The weight of M1, M2, and M3 are 0.25, 0.55 , and 0.2.

\subsubsection{DTIQA}
Team DTIQA wins second place in track 2.
Their main contributions can be divided into three parts.
Firstly, they use a multi-stage Swin Transformer as the baseline and finetune the model on the PIPAL dataset.
Secondly, several training tricks are used to improve the performance.
These tricks include the loss function, data augmentation and test time augmentation.
For the loss function, both ranking loss and regression loss are used.
The regression loss is the Euclidean loss (MSE).
The ranking loss explicitly exploits the relative ranking of image pairs in the PIPAL dataset.
The loss is formulated as:
$$
\mathcal{L}_{rank}=\frac{2}{N}\sum_{i=0}^N e^{\hat{y}^{2i}-\hat{y}^{2i+1}}\mathbb{I}\{y^{2i}<y^{2i+1}\}.
$$
For the training data augmentation, they use (1) random cropping image into patches, (2) random rotation, (3) weighted sampler to ensure that each batch sees a proportional number of mos scores and (4) colorspace augmentation.
For the testing augmentation, they employ (1) random crop, then infer each image 80 times and get a harmonic mean score and (2) resize the image to 256 with diﬀerent interpolation methods.
Finally, they designed a model that combines Transformer and CNN for multiple resolutions.
The framework of the proposed method is shown in \figurename~\ref{fig:5.2.2}.
As input to the Swin Transformer, they used the flattened output of several blocks of EfficientNet. Embedding and projection locations were also added to match the Transformer dimension and expanded with dummy tokens representing aggregated information.

\subsubsection{JMU-CVLab}
Team JMU-CVLab wins third place in track 2.
In their method~\cite{conde2022conformer}, a simple CNN backbone $\phi$ takes a distorted image $x$ as input and aims to minimize the MOS $y$ using the following loss function from \cite{asna2021ntire}, where $\phi(x)=y'$.
$$
\mathcal{L}=\mathrm{MSE}(y,y')+(1-\mathrm{Pearson(y,y')})
$$
They argue that the overfitting problem is the main problem in this track because there are only 200 reference images for 23200 distorted images.
They use several augmentation methods:
Horizontal and vertical flips.
Rotations of 90/180/270 degrees.
Take a random crop of size (224, 224).
One of CutOut or GridMask as further regularization to ensure the model learns to assess the quality without looking at the entire image.
The main trick in their method is called Noisy Student Training, a semi-supervised learning approach that extends the idea of self-training and distillation.
They distinguish a teacher model trained with full-reference pairs and a student model trained only with distorted images.
The process is as follows: (1) train a teacher model using the PIPAL dataset with reference and distorted pairs, (2) infer on unlabeled samples and annotate the images -- these MOS annotations are noisy and called pseudo-labels, (3) add the pseudo-labelled samples to the training set, and (4) train a student model that takes the distorted images as input and trained on the extended datasets.

The organizers note that the new data, which is included in the solution of team JMU-CVLab, contains the validation images from the NTIRE IQA challenge (2021 and 2022).
These data do not contain the final test data, but the distortion type of these data is more similar to the distortion of the test data.
We note that this method has the possibility of not generalising well to other distortion types.

% \subsubsection{KS}
% %
% Team KS designs a Multi-Scale Image Quality Network (MSIQ-Net), which receives a pair of distortion and reference images as input, and generates multi-scale features using a FPN-like module.

\subsubsection{NetEase OPDAI}
Team NetEase OPDAI uses Swin Transformer \cite{liu2021swin} as the backbone.
The Swin Transformer is pretrained on ImageNet~\cite{deng2009imagenet} dataset.
The network can extract more expressive features compared to Resnet~\cite{he2016deep}.
The training uses MSE loss and KL-divergence loss.
They also use the ensemble method to improve the final results.
Two models are used: (1) Swin Transformer Large-224 as the backbone, 2-layer transformer layer to predict MOS score; data augmentation including flipping, cropping, and resizing, and (2) Swin Transformer Tiny-224 as the backbone, 1-layer transformer layer to predict MOS score; data augmentation including flipping, cropping and resizing. In the model (2), they remove the extremely hard distortion type.
The results of these two models are then averaged as the final result.

\subsubsection{Withdrawn submission}
This team withdrew their submission and remains anonymous in this report.
Their method is shown in \figurename~\ref{fig:5.2.5} summarized as follow.
They build an NR-IQA model based on the model structure and loss of Transformers, Relative Ranking, and Self-Consistency (TReS) \cite{golestaneh2022no}.
They use a pre-trained ResNeXt 101 model \cite{xie2017aggregated} on ImageNet \cite{deng2009imagenet} as a feature extractor and fixed it during the IQA training process.
They use fast fourier convolution (FFC) \cite{suvorov2022resolution,chi2020fast} so that the model can employ both global and local contexts of the image for quality assessment.
FFC uses spectral transform and 2D convolutions to increase the receptive field size and learn the global context.
The feature map of each level calculated by the feature extractor goes through normalization and FFC blocks.
Then, these feature maps are combined via pooling and concatenation, passed through a transformer with positional encoding, and then passed through fully-connected layers to be a score value.
For training, they use the L1 loss as the score difference loss when training the model.
In addition, a self-consistency loss is used that allows similar feature vectors to be extracted for the rotated image and a relative ranking loss that considers sample ranking in a mini-batch which are used in TReS as well.
They add an extra loss that narrows the difference between the predicted score difference and the ground truth score difference for every pair in the mini-batch.
An Adam optimizer with a learning rate of $2\times10^{-5}$ is used, and the learning rate is halved after every ten epochs.

\subsubsection{NTU607QCO-IQA}
The method of team NTU607QCO-IQA is also adapted from \cite{golestaneh2022no}.
This model contains a backbone to extract multi-scale features and a linear layer neck to integrate features and output the final scores.
Based on this architecture, they use the Res2Net \cite{gao2019res2net} as the backbone.
Furthermore, in the loss function part, they not only apply the L1 loss but also Pearson’s correlation loss \cite{ayyoubzadeh2021asna} and triplet loss.
The Pearson’s correlation loss is helpful for the model to increase the performance of PLCC.
The triplet loss is used for surrogate ranking.
They use Adamw optimizer with the learning rate of 0.0001 and the learning rate decrease of 0.1 every ten epochs. The total epoch is 50 and takes 11 hours. The batch size is 10.
No data augmentation is used in the training phase.

\section*{Acknowledgments}
We thank the NTIRE 2022 sponsors: Huawei, Reality Labs, Bending Spoons, MediaTek, OPPO, Oddity, Voyage81, ETH Z\"urich (Computer Vision Lab) and University of W\"urzburg (CAIDAS).

\appendix

\section{NTIRE 2022 Organizers}
\noindent\textit{\textbf{Title: }}\\ NTIRE 2022 Challenge on Perceptual Image Quality Assessment\\
\noindent\textit{\textbf{Members:}}\\ \textit{Jinjin Gu$^{1,2}$ (jinjin.gu@sydney.edu.au)}, Haoming Cai$^3$, Chao Dong$^{1,3}$,  Jimmy S. Ren$^4$, Radu Timofte$^{5,6}$\\
\noindent\textit{\textbf{Affiliations: }}\\
$^1$ Shanghai AI Lab, Shanghai, China\\
$^2$ School of Electrical and Information Engineering, The University of Sydney\\
$^3$ Shenzhen Institutes of Advanced Technology, Chinese Academy of Sciences\\
$^4$ SenseTime Research\\
$^5$ University of W\"urzburg, Germany\\
$^6$ ETH Z\"urich, Switzerland\\

\section{Track 1: Teams and Affiliations}
\label{sec:apd:track1team}

\subsection*{THU1919Group}
\noindent\textit{\textbf{Title:}}\\
Attention Helps CNN See Better: Hybrid Image Quality Assessment Network \cite{2022Attentions}\\
\noindent\textit{\textbf{Members: }}\\
\textit{Yuan Gong$^1$ (gong-y21@mails.tsinghua.edu.cn)}, \textit{Shanshan Lao$^1$ (laoss21@mails.tsinghua.edu.cn)}, Shuwei Shi$^1$, Jiahao Wang$^2$, Sidi Yang$^1$, Tianhe Wu$^1$, Weihao Xia$^3$, Yujiu Yang$^1$\\
\noindent\textit{\textbf{Affiliations: }}\\
$^1$ Tsinghua Shenzhen International Graduate School, Tsinghua University\\
$^2$ Department of Automation, Tsinghua University\\
$^3$ University College London\\

\subsection*{Netease OPDAI}
\noindent\textit{\textbf{Title:}}\\
An Algorithm for Reference Image Quality Assessment \cite{2022Image}\\
\noindent\textit{\textbf{Members: }}\\
\textit{Cong Heng$^1$ (congheng@corp.netease.com)}, Lingzhi Fu$^1$, Rongyu Zhang$^1$, Yusheng Zhang$^1$, Hao Wang$^1$, Hongjian Song$^1$\\
\noindent\textit{\textbf{Affiliations: }}\\
$^1$ Netease\\

\subsection*{KS}
\noindent\textit{\textbf{Title:}}\\
Bag of Tricks for Practical Image Quality Assessment with Newly-appeared Distortions\\
\noindent\textit{\textbf{Members: }}\\
\textit{Ming Sun$^1$ (sunming03@kuaishou.com)}, Mading Li$^1$, Kai Zhao$^1$, Kun Yuan$^1$, Zishang Kong$^1$, Mingda Wu$^1$, Chuanchuan Zheng$^1$\\
\noindent\textit{\textbf{Affiliations: }}\\
$^1$ Kuaishou\\

\subsection*{JMU-CVLab}
\noindent\textit{\textbf{Title:}}\\
IQA Conformer Network \cite{2022Conformer}\\
\noindent\textit{\textbf{Members: }}\\
\textit{Maxime Burchi$^1$ (marcos.conde-osorio@uni-wuerzburg.de)}, Marcos V. Conde$^1$, Radu Timofte$^1$\\
\noindent\textit{\textbf{Affiliations: }}\\
$^1$ University of Wurzburg, Computer Vision Lab\\

\subsection*{Yahaha!}
% \noindent\textit{\textbf{Title:}}\\
\noindent\textit{\textbf{Members: }}\\
\textit{Longtao Feng$^1$ (fenglongtao@bytedance.com)}, Tao Zhang$^1$, Yang Li$^1$, Jingwen Xu$^1$, Haiqiang Wang$^1$, Yiting Liao$^1$, Junlin Li$^1$\\
\noindent\textit{\textbf{Affiliations: }}\\
$^1$ ByteDance\\

\subsection*{debut\_kele}
\noindent\textit{\textbf{Title:}}\\
Gradient Boosting Trees -Based Perceptual Image Quality Assessment\\
\noindent\textit{\textbf{Members: }}\\
\textit{Kele Xu$^1$ (kelele.xu@gmail.com)}, Tao Sun$^1$, Yunsheng Xiong$^1$\\
\noindent\textit{\textbf{Affiliations: }}\\
$^1$ Key Laboratory for Parallel and Distributed Processing\\

\subsection*{Pico Zen}
\noindent\textit{\textbf{Title:}}\\
Multi Scale Image Quality Assessment with Transformers by weight sharing approach \cite{2022multi}\\
\noindent\textit{\textbf{Members: }}\\
\textit{Abhisek Keshari$^1$ (2018ume0126@iitjammu.ac.in)}, Komal$^1$, Sadbhawana Thakur$^1$, Vinit Jakhetiya$^1$, Badri N Subudhi$^1$\\
\noindent\textit{\textbf{Affiliations: }}\\
$^1$ Indian Institute of Technology Jammu\\

\subsection*{Team Horizon}
\noindent\textit{\textbf{Title:}}\\
Multi-branch Image Quality Assessment Network \cite{2022can}\\
\noindent\textit{\textbf{Members: }}\\
\textit{Saikat Dutta$^1$ (saikat.dutta779@gmail.com)}, Sourya Dipta Das$^2$, Nisarg A. Shah$^3$, Anil Kumar Tiwari$^3$\\
\noindent\textit{\textbf{Affiliations: }}\\
$^1$ Indian Institute of Technology Madras\\
$^2$ Jadavpur University\\
$^3$ Indian Institute of Technology Jodhpur\\

\section{Track 2: Teams and Affiliations}
\label{sec:apd:track2team}

\subsection*{THU\_IIGROUP}
\noindent\textit{\textbf{Title:}}\\
Multi-Dimension Attention Network for Image Quality Assessment \cite{2022MANIQA}\\
\noindent\textit{\textbf{Members: }}\\
\textit{Sidi Yang$^1$ (yangsd21@mails.tsinghua.edu.cn)}, \textit{Tianhe Wu$^1$ (tianhe wu@foxmail.com)}, Shuwei Shi$^1$, Shanshan Lao$^1$, Yuan Gong$^1$, Mingdeng Cao$^1$, Jiahao Wang$^1$, Yujiu Yang$^1$\\
\noindent\textit{\textbf{Affiliations: }}\\
$^1$ Tsinghua Shenzhen International Graduate School, Tsinghua University\\
$^2$ Department of Automation, Tsinghua University\\

\subsection*{DTIQA}
\noindent\textit{\textbf{Title:}}\\
Assessment Method Based on Transformer and Convolutional Neural Network\\
\noindent\textit{\textbf{Members: }}\\
\textit{Jing Wang$^1$ (wangjing.crystalw@bytedance.com)}, Haotian Fan$^1$, Xiaoxia Hou$^1$\\
\noindent\textit{\textbf{Affiliations: }}\\
$^1$ ByteDance\\

\subsection*{JMU-CVLab}
\noindent\textit{\textbf{Title:}}\\
Blind and Noisy IQA Students \cite{2022Conformer}\\
\noindent\textit{\textbf{Members: }}\\
\textit{Marcos V. Conde$^1$ (marcos.conde-osorio@uni-wuerzburg.de)}, Maxime Burchi$^1$, Radu Timofte$^1$\\
\noindent\textit{\textbf{Affiliations: }}\\
$^1$ University of W\"urzburg, Computer Vision Lab\\

\subsection*{KS}
\noindent\textit{\textbf{Title:}}\\
Bag of Tricks for Practical Image Quality Assessment with Newly-appeared Distortions\\
\noindent\textit{\textbf{Members: }}\\
\textit{Ming Sun$^1$ (sunming03@kuaishou.com)}, Mading Li$^1$, Kai Zhao$^1$, Kun Yuan$^1$, Zishang Kong$^1$, Mingda Wu$^1$, Chuanchuan Zheng$^1$\\
\noindent\textit{\textbf{Affiliations: }}\\
$^1$ Kuaishou\\

\subsection*{NetEase OPDAI}
% \noindent\textit{\textbf{Title:}}\\
\noindent\textit{\textbf{Members: }}\\
\textit{Heng Cong$^1$ (congheng@corp.netease.com)}, Lingzhi Fu$^1$, Rongyu Zhang$^1$, Yusheng Zhang$^1$, Hao Wang$^1$, Hongjian Song$^1$\\
\noindent\textit{\textbf{Affiliations: }}\\
$^1$ Netease\\

\subsection*{NTU607QCO-IQA}
\noindent\textit{\textbf{Title:}}\\
Multiple loss functions for no reference perceptual image quality assessment\\
\noindent\textit{\textbf{Members: }}\\
\textit{Hao-Hsiang Yang$^1$ (islike8399@gmail.com)}, Hua-En Chang$^1$, Zhi-Kai Huang$^1$, Wei-Ting Chen$^1$, Sy-Yen Kuo$^1$\\
\noindent\textit{\textbf{Affiliations: }}\\
$^1$ National Taiwan University\\

% \section{Change Log}
% \label{sec:apd:change}

{\small
\bibliographystyle{ieee_fullname}
\bibliography{egbib}

\begin{thebibliography}{10}\itemsep=-1pt

\bibitem{arad2022ntiredemosaicing}
Boaz Arad, Radu Timofte, Rony Yahel, Nimrod Morag, Amir Bernat, et~al.
\newblock {NTIRE} 2022 spectral demosaicing challenge and dataset.
\newblock In {\em Proceedings of the IEEE/CVF Conference on Computer Vision and
  Pattern Recognition (CVPR) Workshops}, 2022.

\bibitem{arad2022ntirerecovery}
Boaz Arad, Radu Timofte, Rony Yahel, Nimrod Morag, Amir Bernat, et~al.
\newblock {NTIRE} 2022 spectral recovery challenge and dataset.
\newblock In {\em Proceedings of the IEEE/CVF Conference on Computer Vision and
  Pattern Recognition (CVPR) Workshops}, 2022.

\bibitem{ayyoubzadeh2021asna}
Seyed~Mehdi Ayyoubzadeh and Ali Royat.
\newblock (asna) an attention-based siamese-difference neural network with
  surrogate ranking loss function for perceptual image quality assessment.
\newblock In {\em Proceedings of the IEEE/CVF Conference on Computer Vision and
  Pattern Recognition}, pages 388--397, 2021.

\bibitem{asna2021ntire}
Seyed~Mehdi Ayyoubzadeh and Ali Royat.
\newblock (asna) an attention-based siamese-difference neural network with
  surrogate ranking loss function for perceptual image quality assessment.
\newblock In {\em IEEE/CVF Conference on Computer Vision and Pattern
  Recognition Workshops}, 2021.

\bibitem{benesty2009pearson}
Jacob Benesty, Jingdong Chen, Yiteng Huang, and Israel Cohen.
\newblock Pearson correlation coefficient.
\newblock In {\em Noise reduction in speech processing}, pages 1--4. Springer,
  2009.

\bibitem{bhat2022ntire}
Goutam Bhat, Martin Danelljan, Radu Timofte, et~al.
\newblock {NTIRE} 2022 burst super-resolution challenge.
\newblock In {\em Proceedings of the IEEE/CVF Conference on Computer Vision and
  Pattern Recognition (CVPR) Workshops}, 2022.

\bibitem{bianco2018use}
Simone Bianco, Luigi Celona, Paolo Napoletano, and Raimondo Schettini.
\newblock On the use of deep learning for blind image quality assessment.
\newblock {\em Signal, Image and Video Processing}, 12(2):355--362, 2018.

\bibitem{blau2018perception}
Yochai Blau and Tomer Michaeli.
\newblock The perception-distortion tradeoff.
\newblock In {\em Proceedings of the IEEE Conference on Computer Vision and
  Pattern Recognition}, pages 6228--6237, 2018.

\bibitem{wadiqam}
Sebastian Bosse, Dominique Maniry, Klaus-Robert M{\"u}ller, Thomas Wiegand, and
  Wojciech Samek.
\newblock Deep neural networks for no-reference and full-reference image
  quality assessment.
\newblock {\em IEEE Transactions on Image Processing}, 27(1):206--219, 2017.

\bibitem{bosse2017deep}
Sebastian Bosse, Dominique Maniry, Klaus-Robert M{\"u}ller, Thomas Wiegand, and
  Wojciech Samek.
\newblock Deep neural networks for no-reference and full-reference image
  quality assessment.
\newblock {\em IEEE Transactions on image processing}, 27(1):206--219, 2017.

\bibitem{burchi2021efficient}
Maxime Burchi and Valentin Vielzeuf.
\newblock Efficient conformer: Progressive downsampling and grouped attention
  for automatic speech recognition.
\newblock {\em arXiv preprint arXiv:2109.01163}, 2021.

\bibitem{Cai2021CUGAN}
Haoming Cai, Jingwen He, Yu Qiao, and Chao Dong.
\newblock Toward interactive modulation for photo-realistic image restoration.
\newblock In {\em Proceedings of the IEEE/CVF Conference on Computer Vision and
  Pattern Recognition (CVPR) Workshops}, pages 294--303, June 2021.

\bibitem{IQT2021ntire}
Manri Cheon, Sung-Jun Yoon, Byungyeon Kang, and Junwoo Lee.
\newblock Perceptual image quality assessment with transformers.
\newblock In {\em IEEE/CVF Conference on Computer Vision and Pattern
  Recognition Workshops}, 2021.

\bibitem{chi2020fast}
Lu Chi, Borui Jiang, and Yadong Mu.
\newblock Fast fourier convolution.
\newblock {\em Advances in Neural Information Processing Systems},
  33:4479--4488, 2020.

\bibitem{conde2022conformer}
Marcos~V. Conde, Maxime Burchi, and Radu Timofte.
\newblock Conformer and blind noisy students for improved image quality
  assessment.
\newblock In {\em Proceedings of the IEEE/CVF Conference on Computer Vision and
  Pattern Recognition (CVPR) Workshops}, 2022.

\bibitem{2022Conformer}
Marcos~V Conde, Maxime Burchi, and Radu Timofte.
\newblock Conformer and blind noisy students for improved image quality
  assessment.
\newblock In {\em IEEE/CVF Conference on Computer Vision and Pattern
  Recognition Workshops}, 2022.

\bibitem{2022Image}
Heng Cong, Lingzhi Fu, Rongyu Zhang, Yusheng Zhang, Hao Wang, Jiarong He, and
  Jin Gao.
\newblock Image quality assessment with gradient siamese network.
\newblock In {\em IEEE/CVF Conference on Computer Vision and Pattern
  Recognition Workshops}, 2022.

\bibitem{deng2009imagenet}
Jia Deng, Wei Dong, Richard Socher, Li-Jia Li, Kai Li, and Li Fei-Fei.
\newblock Imagenet: A large-scale hierarchical image database.
\newblock In {\em 2009 IEEE conference on computer vision and pattern
  recognition}, pages 248--255. Ieee, 2009.

\bibitem{dists}
Keyan Ding, Kede Ma, Shiqi Wang, and Eero~P Simoncelli.
\newblock Image quality assessment: Unifying structure and texture similarity.
\newblock {\em IEEE Transactions on Pattern Analysis and Machine Intelligence},
  2020.

\bibitem{dosovitskiy2020image}
Alexey Dosovitskiy, Lucas Beyer, Alexander Kolesnikov, Dirk Weissenborn,
  Xiaohua Zhai, Thomas Unterthiner, Mostafa Dehghani, Matthias Minderer, Georg
  Heigold, Sylvain Gelly, et~al.
\newblock An image is worth 16x16 words: Transformers for image recognition at
  scale.
\newblock {\em arXiv preprint arXiv:2010.11929}, 2020.

\bibitem{2022can}
Saikat Dutta, Sourya~Dipta Das, and Nisarg~A Shah.
\newblock Can no-reference features help in full-reference image quality
  estimation?
\newblock {\em arXiv preprint arXiv:2203.00845}, 2022.

\bibitem{elo1978rating}
Arpad~E Elo.
\newblock {\em The rating of chessplayers, past and present}.
\newblock Arco Pub., 1978.

\bibitem{ershov2022ntire}
Egor Ershov, Alex Savchik, Denis Shepelev, Nikola Banic, Michael~S Brown, Radu
  Timofte, et~al.
\newblock {NTIRE} 2022 challenge on night photography rendering.
\newblock In {\em Proceedings of the IEEE/CVF Conference on Computer Vision and
  Pattern Recognition (CVPR) Workshops}, 2022.

\bibitem{fang2020perceptual}
Yuming Fang, Hanwei Zhu, Yan Zeng, Kede Ma, and Zhou Wang.
\newblock Perceptual quality assessment of smartphone photography.
\newblock In {\em Proceedings of the IEEE/CVF Conference on Computer Vision and
  Pattern Recognition}, pages 3677--3686, 2020.

\bibitem{gao2019res2net}
Shang-Hua Gao, Ming-Ming Cheng, Kai Zhao, Xin-Yu Zhang, Ming-Hsuan Yang, and
  Philip Torr.
\newblock Res2net: A new multi-scale backbone architecture.
\newblock {\em IEEE transactions on pattern analysis and machine intelligence},
  43(2):652--662, 2019.

\bibitem{golestaneh2022no}
S~Alireza Golestaneh, Saba Dadsetan, and Kris~M Kitani.
\newblock No-reference image quality assessment via transformers, relative
  ranking, and self-consistency.
\newblock In {\em Proceedings of the IEEE/CVF Winter Conference on Applications
  of Computer Vision}, pages 1220--1230, 2022.

\bibitem{goodfellow2014generative}
Ian Goodfellow, Jean Pouget-Abadie, Mehdi Mirza, Bing Xu, David Warde-Farley,
  Sherjil Ozair, Aaron Courville, and Yoshua Bengio.
\newblock Generative adversarial nets.
\newblock In {\em Advances in neural information processing systems}, pages
  2672--2680, 2014.

\bibitem{gu2020image}
Jinjin Gu, Haoming Cai, Haoyu Chen, Xiaoxing Ye, Jimmy Ren, and Chao Dong.
\newblock Image quality assessment for perceptual image restoration: A new
  dataset, benchmark and metric.
\newblock {\em arXiv preprint arXiv:2011.15002}, 2020.

\bibitem{pipal}
Jinjin Gu, Haoming Cai, Haoyu Chen, Xiaoxing Ye, Jimmy~S Ren, and Chao Dong.
\newblock Pipal: a large-scale image quality assessment dataset for perceptual
  image restoration.
\newblock In {\em Proceedings of the European Conference on Computer Vision
  (ECCV)}, 2020.

\bibitem{gu2022ntire}
Jinjin Gu, Haoming Cai, Chao Dong, Jimmy Ren, Radu Timofte, et~al.
\newblock {NTIRE} 2022 challenge on perceptual image quality assessment.
\newblock In {\em Proceedings of the IEEE/CVF Conference on Computer Vision and
  Pattern Recognition (CVPR) Workshops}, 2022.

\bibitem{gu2021ntire}
Jinjin Gu, Haoming Cai, Chao Dong, Jimmy~S Ren, Yu Qiao, Shuhang Gu, and Radu
  Timofte.
\newblock Ntire 2021 challenge on perceptual image quality assessment.
\newblock In {\em Proceedings of the IEEE/CVF Conference on Computer Vision and
  Pattern Recognition}, pages 677--690, 2021.

\bibitem{gulati2020conformer}
Anmol Gulati, James Qin, Chung-Cheng Chiu, Niki Parmar, Yu Zhang, Jiahui Yu,
  Wei Han, Shibo Wang, Zhengdong Zhang, Yonghui Wu, et~al.
\newblock Conformer: Convolution-augmented transformer for speech recognition.
\newblock {\em arXiv preprint arXiv:2005.08100}, 2020.

\bibitem{IQMA2021ntire}
Haiyang Guo, Yi Bin, Yuqing Hou, Qing Zhang, and Hengliang Luo.
\newblock Iqma network: Image quality multi-scale assessment network.
\newblock In {\em IEEE/CVF Conference on Computer Vision and Pattern
  Recognition Workshops}, 2021.

\bibitem{he2016deep}
Kaiming He, Xiangyu Zhang, Shaoqing Ren, and Jian Sun.
\newblock Deep residual learning for image recognition.
\newblock In {\em Proceedings of the IEEE conference on computer vision and
  pattern recognition}, pages 770--778, 2016.

\bibitem{johnson2016perceptual}
Justin Johnson, Alexandre Alahi, and Li Fei-Fei.
\newblock Perceptual losses for real-time style transfer and super-resolution.
\newblock In {\em European conference on computer vision}, pages 694--711.
  Springer, 2016.

\bibitem{kang2014convolutional}
Le Kang, Peng Ye, Yi Li, and David Doermann.
\newblock Convolutional neural networks for no-reference image quality
  assessment.
\newblock In {\em Proceedings of the IEEE conference on computer vision and
  pattern recognition}, pages 1733--1740, 2014.

\bibitem{2022multi}
Abhisek Keshari, Komal, Sadbhawna, and Badri Subudhi.
\newblock Multi-scale features and parallel transformers based image quality
  assessment.
\newblock {\em arXiv preprint arXiv:2204.09779}, 2022.

\bibitem{2022Attentions}
Shanshan Lao, Yuan Gong, Shuwei Shi, Sidi Yang, Tianhe Wu, Jiahao Wang, Weihao
  Xia, and Yujiu Yang.
\newblock Attentions help cnns see better: Attention-based hybrid image quality
  assessment network.
\newblock In {\em IEEE/CVF Conference on Computer Vision and Pattern
  Recognition Workshops}, 2022.

\bibitem{srgan2017}
Christian Ledig, Lucas Theis, Ferenc Husz{\'a}r, Jose Caballero, Andrew
  Cunningham, Alejandro Acosta, Andrew Aitken, Alykhan Tejani, Johannes Totz,
  Zehan Wang, et~al.
\newblock Photo-realistic single image super-resolution using a generative
  adversarial network.
\newblock In {\em Proceedings of the IEEE conference on computer vision and
  pattern recognition}, pages 4681--4690, 2017.

\bibitem{li2020norm}
Dingquan Li, Tingting Jiang, and Ming Jiang.
\newblock Norm-in-norm loss with faster convergence and better performance for
  image quality assessment.
\newblock In {\em Proceedings of the 28th ACM International Conference on
  Multimedia}, pages 789--797, 2020.

\bibitem{li2021reproducibility}
Dingquan Li, Tingting Jiang, Ming Jiang, Vajira~Lasantha Thambawita, and
  Haoliang Wang.
\newblock Reproducibility companion paper: Norm-in-norm loss with faster
  convergence and better performance for image quality assessment.
\newblock In {\em Proceedings of the 29th ACM International Conference on
  Multimedia}, pages 3615--3618, 2021.

\bibitem{li2022ntire}
Yawei Li, Kai Zhang, Radu Timofte, Luc Van~Gool, et~al.
\newblock {NTIRE} 2022 challenge on efficient super-resolution: Methods and
  results.
\newblock In {\em Proceedings of the IEEE/CVF Conference on Computer Vision and
  Pattern Recognition (CVPR) Workshops}, 2022.

\bibitem{lin2018hallucinated}
Kwan-Yee Lin and Guanxiang Wang.
\newblock Hallucinated-iqa: No-reference image quality assessment via
  adversarial learning.
\newblock In {\em Proceedings of the IEEE Conference on Computer Vision and
  Pattern Recognition}, pages 732--741, 2018.

\bibitem{liu2021swin}
Ze Liu, Yutong Lin, Yue Cao, Han Hu, Yixuan Wei, Zheng Zhang, Stephen Lin, and
  Baining Guo.
\newblock Swin transformer: Hierarchical vision transformer using shifted
  windows.
\newblock In {\em Proceedings of the IEEE/CVF International Conference on
  Computer Vision}, pages 10012--10022, 2021.

\bibitem{loshchilov2017decoupled}
Ilya Loshchilov and Frank Hutter.
\newblock Decoupled weight decay regularization.
\newblock {\em arXiv preprint arXiv:1711.05101}, 2017.

\bibitem{lugmayr2022ntire}
Andreas Lugmayr, Martin Danelljan, Radu Timofte, et~al.
\newblock {NTIRE} 2022 challenge on learning the super-resolution space.
\newblock In {\em Proceedings of the IEEE/CVF Conference on Computer Vision and
  Pattern Recognition (CVPR) Workshops}, 2022.

\bibitem{ma2017learning}
Chao Ma, Chih-Yuan Yang, Xiaokang Yang, and Ming-Hsuan Yang.
\newblock Learning a no-reference quality metric for single-image
  super-resolution.
\newblock {\em Computer Vision and Image Understanding}, 158:1--16, 2017.

\bibitem{mittal2012making}
Anish Mittal, Rajiv Soundararajan, and Alan~C Bovik.
\newblock Making a “completely blind” image quality analyzer.
\newblock {\em IEEE Signal processing letters}, 20(3):209--212, 2012.

\bibitem{niqe}
Anish Mittal, Rajiv Soundararajan, and Alan~C Bovik.
\newblock Making a “completely blind” image quality analyzer.
\newblock {\em IEEE Signal Processing Letters}, 20(3):209--212, 2012.

\bibitem{moorthy2011blind}
Anush~Krishna Moorthy and Alan~Conrad Bovik.
\newblock Blind image quality assessment: From natural scene statistics to
  perceptual quality.
\newblock {\em IEEE transactions on Image Processing}, 20(12):3350--3364, 2011.

\bibitem{pan2022no}
Zhaoqing Pan, Feng Yuan, Xu Wang, Long Xu, Shao Xiao, and Sam Kwong.
\newblock No-reference image quality assessment via multi-branch convolutional
  neural networks.
\newblock {\em IEEE Transactions on Artificial Intelligence}, 2022.

\bibitem{perezpellitero2022ntire}
Eduardo P\'erez-Pellitero, Sibi Catley-Chandar, Richard Shaw, Ales Leonardis,
  Radu Timofte, et~al.
\newblock {NTIRE} 2022 challenge on high dynamic range imaging: Methods and
  results.
\newblock In {\em Proceedings of the IEEE/CVF Conference on Computer Vision and
  Pattern Recognition (CVPR) Workshops}, 2022.

\bibitem{tid2013}
Nikolay Ponomarenko, Lina Jin, Oleg Ieremeiev, Vladimir Lukin, Karen
  Egiazarian, Jaakko Astola, Benoit Vozel, Kacem Chehdi, Marco Carli, Federica
  Battisti, et~al.
\newblock Image database tid2013: Peculiarities, results and perspectives.
\newblock {\em Signal Processing: Image Communication}, 30:57--77, 2015.

\bibitem{prashnani2018pieapp}
Ekta Prashnani, Hong Cai, Yasamin Mostofi, and Pradeep Sen.
\newblock Pieapp: Perceptual image-error assessment through pairwise
  preference.
\newblock In {\em Proceedings of the IEEE Conference on Computer Vision and
  Pattern Recognition}, pages 1808--1817, 2018.

\bibitem{romero2022ntire}
Andres Romero, Angela Castillo, Jose~M Abril-Nova, Radu Timofte, et~al.
\newblock {NTIRE} 2022 image inpainting challenge: Report.
\newblock In {\em Proceedings of the IEEE/CVF Conference on Computer Vision and
  Pattern Recognition (CVPR) Workshops}, 2022.

\bibitem{saad2012blind}
Michele~A Saad, Alan~C Bovik, and Christophe Charrier.
\newblock Blind image quality assessment: A natural scene statistics approach
  in the dct domain.
\newblock {\em IEEE transactions on Image Processing}, 21(8):3339--3352, 2012.

\bibitem{ifc}
Hamid~R Sheikh, Alan~C Bovik, and Gustavo De~Veciana.
\newblock An information fidelity criterion for image quality assessment using
  natural scene statistics.
\newblock {\em IEEE Transactions on image processing}, 14(12):2117--2128, 2005.

\bibitem{sheikh2006statistical}
Hamid~R Sheikh, Muhammad~F Sabir, and Alan~C Bovik.
\newblock A statistical evaluation of recent full reference image quality
  assessment algorithms.
\newblock {\em IEEE Transactions on image processing}, 15(11):3440--3451, 2006.

\bibitem{RADN2021ntire}
Shuwei Shi, Qingyan Bai, Mingdeng Cao, Weihao Xia, Jiahao Wang, Yifan Chen, and
  Yujiu Yang.
\newblock Region-adaptive deformable network for image quality assessment.
\newblock In {\em IEEE/CVF Conference on Computer Vision and Pattern
  Recognition Workshops}, 2021.

\bibitem{su2020blindly}
Shaolin Su, Qingsen Yan, Yu Zhu, Cheng Zhang, Xin Ge, Jinqiu Sun, and Yanning
  Zhang.
\newblock Blindly assess image quality in the wild guided by a self-adaptive
  hyper network.
\newblock In {\em Proceedings of the IEEE/CVF Conference on Computer Vision and
  Pattern Recognition}, pages 3667--3676, 2020.

\bibitem{suvorov2022resolution}
Roman Suvorov, Elizaveta Logacheva, Anton Mashikhin, Anastasia Remizova,
  Arsenii Ashukha, Aleksei Silvestrov, Naejin Kong, Harshith Goka, Kiwoong
  Park, and Victor Lempitsky.
\newblock Resolution-robust large mask inpainting with fourier convolutions.
\newblock In {\em Proceedings of the IEEE/CVF Winter Conference on Applications
  of Computer Vision}, pages 2149--2159, 2022.

\bibitem{szegedy2017inception}
Christian Szegedy, Sergey Ioffe, Vincent Vanhoucke, and Alexander Alemi.
\newblock Inception-v4, inception-resnet and the impact of residual connections
  on learning.
\newblock In {\em Proceedings of the AAAI Conference on Artificial
  Intelligence}, volume~31, 2017.

\bibitem{talebi2018nima}
Hossein Talebi and Peyman Milanfar.
\newblock Nima: Neural image assessment.
\newblock {\em IEEE transactions on image processing}, 27(8):3998--4011, 2018.

\bibitem{wang2022ntire}
Longguang Wang, Yulan Guo, Yingqian Wang, Juncheng Li, Shuhang Gu, Radu
  Timofte, et~al.
\newblock {NTIRE} 2022 challenge on stereo image super-resolution: Methods and
  results.
\newblock In {\em Proceedings of the IEEE/CVF Conference on Computer Vision and
  Pattern Recognition (CVPR) Workshops}, 2022.

\bibitem{sftgan2018}
Xintao Wang, Ke Yu, Chao Dong, and Chen Change~Loy.
\newblock Recovering realistic texture in image super-resolution by deep
  spatial feature transform.
\newblock In {\em Proceedings of the IEEE Conference on Computer Vision and
  Pattern Recognition}, pages 606--615, 2018.

\bibitem{wang2018esrgan}
Xintao Wang, Ke Yu, Shixiang Wu, Jinjin Gu, Yihao Liu, Chao Dong, Yu Qiao, and
  Chen~Change Loy.
\newblock Esrgan: Enhanced super-resolution generative adversarial networks.
\newblock In {\em European Conference on Computer Vision}, pages 63--79.
  Springer, 2018.

\bibitem{ssim}
Zhou Wang, Alan~C Bovik, Hamid~R Sheikh, Eero~P Simoncelli, et~al.
\newblock Image quality assessment: from error visibility to structural
  similarity.
\newblock {\em IEEE transactions on image processing}, 13(4):600--612, 2004.

\bibitem{ms-ssim}
Zhou Wang, Eero~P Simoncelli, and Alan~C Bovik.
\newblock Multiscale structural similarity for image quality assessment.
\newblock In {\em The Thrity-Seventh Asilomar Conference on Signals, Systems \&
  Computers, 2003}, volume~2, pages 1398--1402, 2003.

\bibitem{xie2017aggregated}
Saining Xie, Ross Girshick, Piotr Doll{\'a}r, Zhuowen Tu, and Kaiming He.
\newblock Aggregated residual transformations for deep neural networks.
\newblock In {\em Proceedings of the IEEE conference on computer vision and
  pattern recognition}, pages 1492--1500, 2017.

\bibitem{yan2019naturalness}
Bo Yan, Bahetiyaer Bare, and Weimin Tan.
\newblock Naturalness-aware deep no-reference image quality assessment.
\newblock {\em IEEE Transactions on Multimedia}, 21(10):2603--2615, 2019.

\bibitem{yang2022ntire}
Ren Yang, Radu Timofte, et~al.
\newblock {NTIRE} 2022 challenge on super-resolution and quality enhancement of
  compressed video: Dataset, methods and results.
\newblock In {\em Proceedings of the IEEE/CVF Conference on Computer Vision and
  Pattern Recognition (CVPR) Workshops}, 2022.

\bibitem{2022MANIQA}
Sidi Yang, Tianhe Wu, Shuwei Shi, Shanshan Lao, Yuan Gong, Cao Mingdeng, Jiahao
  Wang, and Yujiu Yang.
\newblock Maniqa: Multi-dimension attention network for no-reference image
  quality assessment.
\newblock In {\em IEEE/CVF Conference on Computer Vision and Pattern
  Recognition Workshops}, 2022.

\bibitem{ye2012unsupervised}
Peng Ye, Jayant Kumar, Le Kang, and David Doermann.
\newblock Unsupervised feature learning framework for no-reference image
  quality assessment.
\newblock In {\em 2012 IEEE conference on computer vision and pattern
  recognition}, pages 1098--1105. IEEE, 2012.

\bibitem{zamir2021restormer}
Syed~Waqas Zamir, Aditya Arora, Salman Khan, Munawar Hayat, Fahad~Shahbaz Khan,
  and Ming-Hsuan Yang.
\newblock Restormer: Efficient transformer for high-resolution image
  restoration.
\newblock {\em arXiv preprint arXiv:2111.09881}, 2021.

\bibitem{sr-sim}
Lin Zhang and Hongyu Li.
\newblock Sr-sim: A fast and high performance iqa index based on spectral
  residual.
\newblock In {\em 2012 19th IEEE international conference on image processing},
  pages 1473--1476. IEEE, 2012.

\bibitem{vsi}
Lin Zhang, Ying Shen, and Hongyu Li.
\newblock Vsi: A visual saliency-induced index for perceptual image quality
  assessment.
\newblock {\em IEEE Transactions on Image Processing}, 23(10):4270--4281, 2014.

\bibitem{zhang2015feature}
Lin Zhang, Lei Zhang, and Alan~C Bovik.
\newblock A feature-enriched completely blind image quality evaluator.
\newblock {\em IEEE Transactions on Image Processing}, 24(8):2579--2591, 2015.

\bibitem{fsim}
Lin Zhang, Lei Zhang, Xuanqin Mou, and David Zhang.
\newblock Fsim: A feature similarity index for image quality assessment.
\newblock {\em IEEE transactions on Image Processing}, 20(8):2378--2386, 2011.

\bibitem{zhang2015som}
Peng Zhang, Wengang Zhou, Lei Wu, and Houqiang Li.
\newblock Som: Semantic obviousness metric for image quality assessment.
\newblock In {\em Proceedings of the IEEE Conference on Computer Vision and
  Pattern Recognition}, pages 2394--2402, 2015.

\bibitem{zhang2018unreasonable}
Richard Zhang, Phillip Isola, Alexei~A Efros, Eli Shechtman, and Oliver Wang.
\newblock The unreasonable effectiveness of deep features as a perceptual
  metric.
\newblock In {\em Proceedings of the IEEE Conference on Computer Vision and
  Pattern Recognition}, pages 586--595, 2018.

\bibitem{zhang2019ranksrgan}
Wenlong Zhang, Yihao Liu, Chao Dong, and Yu Qiao.
\newblock Ranksrgan: Generative adversarial networks with ranker for image
  super-resolution.
\newblock 2019.

\bibitem{zhang2018blind}
Weixia Zhang, Kede Ma, Jia Yan, Dexiang Deng, and Zhou Wang.
\newblock Blind image quality assessment using a deep bilinear convolutional
  neural network.
\newblock {\em IEEE Transactions on Circuits and Systems for Video Technology},
  30(1):36--47, 2018.

\bibitem{zhang2020learning}
Weixia Zhang, Kede Ma, Guangtao Zhai, and Xiaokang Yang.
\newblock Learning to blindly assess image quality in the laboratory and wild.
\newblock In {\em 2020 IEEE International Conference on Image Processing
  (ICIP)}, pages 111--115. IEEE, 2020.

\bibitem{zhu2020metaiqa}
Hancheng Zhu, Leida Li, Jinjian Wu, Weisheng Dong, and Guangming Shi.
\newblock Metaiqa: Deep meta-learning for no-reference image quality
  assessment.
\newblock In {\em Proceedings of the IEEE/CVF Conference on Computer Vision and
  Pattern Recognition}, pages 14143--14152, 2020.

\end{thebibliography}
}

\end{document}